\documentclass[sn-vancouver,iicol]{sn-jnl}

\usepackage[utf8]{inputenc}
\usepackage{graphicx}
\usepackage{amsmath}
\usepackage{lettrine}
\usepackage{array}
\usepackage{fixltx2e}
\usepackage{stfloats}
\usepackage{amsfonts}
\usepackage{mathtools}
\usepackage{amssymb}
\usepackage{float}
\usepackage{adjustbox}
\usepackage{soul}
\usepackage{xcolor}
\usepackage{natbib}
\usepackage{mathptmx}  
\usepackage{latexsym}
\usepackage{caption}
\usepackage{subcaption}
\jyear{2022}%

\theoremstyle{thmstyleone}%
\theoremstyle{thmstyletwo}%

\theoremstyle{thmstylethree}%

\begin{document}



%





\title[Article Title]{Single-cell Subcellular Protein Localisation Using Novel Ensembles of Diverse Deep Architectures}

\author*[1]{\fnm{Syed Sameed} \sur{Husain}}\email{sameed.husain@surrey.ac.uk}
\author[1]{\fnm{Eng-Jon} \sur{Ong}}\email{e.ong@surrey.ac.uk}
\author[1]{\fnm{Dmitry} \sur{Minskiy}}\email{d.minskiy@surrey.ac.uk}
\author[1,2]{\fnm{Mikel} \sur{Bober-Irizar}}\email{mikel@mxbi.net}
\author[2]{\fnm{Amaia} \sur{Irizar}}\email{a.irizar@forecom.ai}
\author[1,2]{\fnm{Miroslaw} \sur{Bober}}\email{m.bober@surrey.ac.uk}

\affil[1]{\orgdiv{CVSSP}, \orgname{University of Surrey}, \orgaddress{
\city{Guildford}, \postcode{GU27XH}, \state{Surrey}, \country{UK}}}
\affil[2]{\orgname{ForecomAI}, \orgaddress{\city{Leatherhead}, \postcode{KT228QY}, \state{Surrey}, \country{UK}}}


\abstract{
Unravelling protein distributions within individual cells is key to understanding their function and state and indispensable to developing new treatments. Here we present the Hybrid subCellular Protein Localiser (HCPL), which learns from weakly labelled data to robustly localise single-cell subcellular protein patterns. It comprises innovative DNN architectures exploiting wavelet filters and learnt parametric activations that successfully tackle drastic cell variability. HCPL features correlation-based ensembling of novel architectures that boosts performance and aids generalisation. Large-scale data annotation is made feasible by our ''AI-trains-AI'' approach, which determines the visual integrity of cells and emphasises reliable labels for efficient training. In the Human Protein Atlas context, we demonstrate that HCPL defines state-of-the-art in the single-cell classification of protein localisation patterns. To better understand the inner workings of HCPL and assess its biological relevance, we analyse the contributions of each system component and dissect the emergent features from which the localisation predictions are derived.}


\maketitle

Proteins play a vital role in most cellular processes crucial to our survival. Their intracellular locations provide important insights about cell functions and state \cite{Paul}. The specific biological functions that proteins perform are closely tied to the subcellular compartments in which they are expressed. Therefore, the subcellular resolution is critical in determining functional information about proteins and understanding the regulation of individual cells. A highly promising direction in this field is automated analysis of immunofluorescence microscopy images to enable large-scale impactful discoveries.
For example, image-based spatial analysis of proteomic cellular heterogeneity can uncover a valuable view of protein expression with subcellular resolution, aiding the identification of disease biomarkers and drug discovery \cite{GNANN, mahdessian2021spatiotemporal, Chandra}.

Single-cell analysis is key to the detection of rare cells in heterogenous populations, essential in the profiling of tumour biology and precision medicine \cite{tellez2016tumour, bodenmiller2016multiplexed}. It constitutes a core strategy of the LifeTime Initiative, a large-scale, long-term initiative to implement cell-based interceptive medicine in Europe \cite{rajewsky2020lifetime}. While machine learning (ML) has been used to describe the location of human proteins in microscope images giving summary information on an entire \textit{population of cells} \cite{loccat, kagglenature, cytoself}, to the best of our knowledge, no published work exists on the subcellular classification of proteins for \textit{individual cells}.

\begin{figure*}
     \centering
     \vspace{-0.3cm}
     \begin{subfigure}[b]{0.34\textwidth}
         \centering
        \caption{}
         \includegraphics[width=\textwidth]{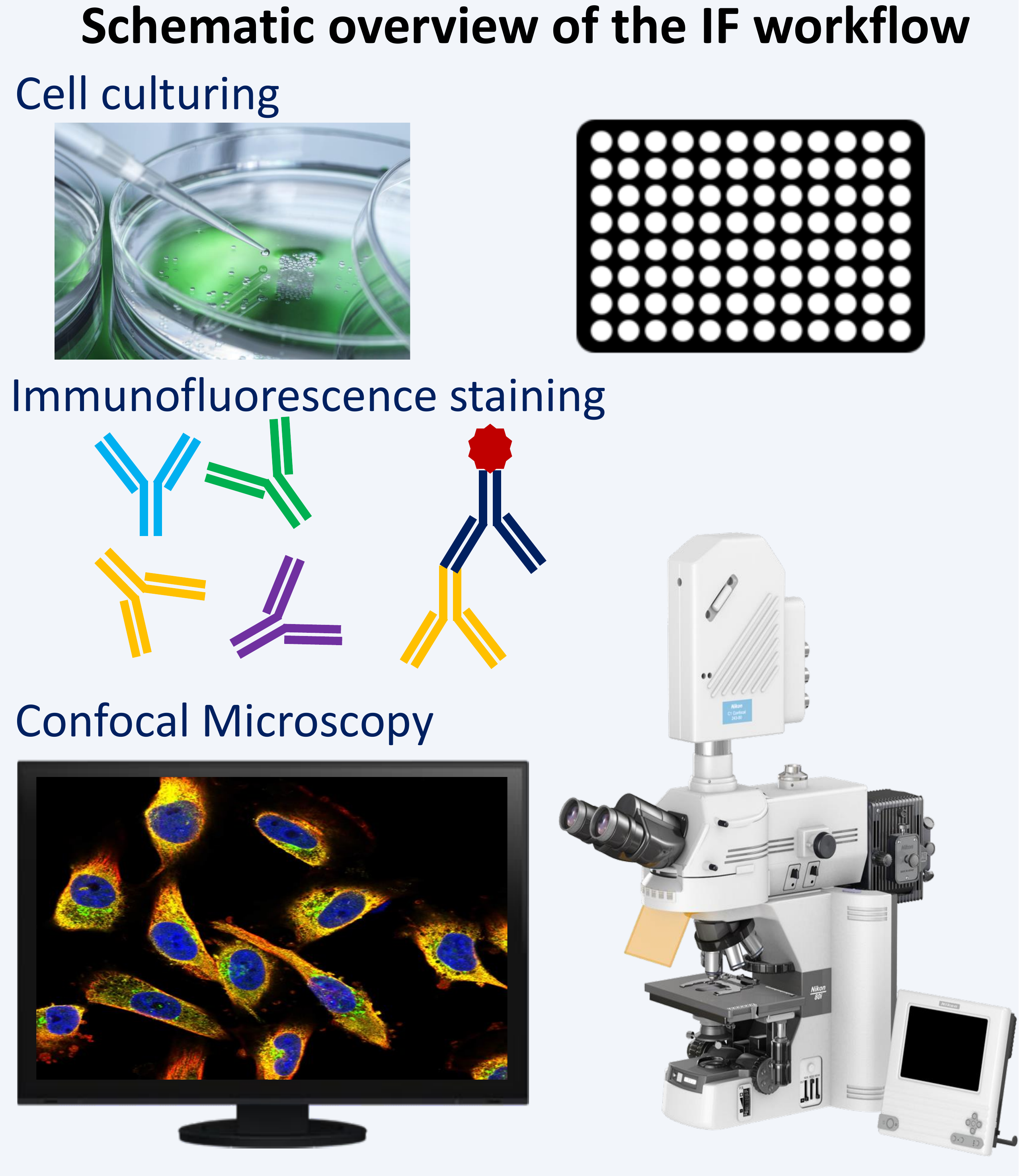}
         \label{fig:confocal}
     \end{subfigure}
     \hfill
     \begin{subfigure}[b]{0.26\textwidth}
         \centering
        \caption{}
         \includegraphics[width=\textwidth]{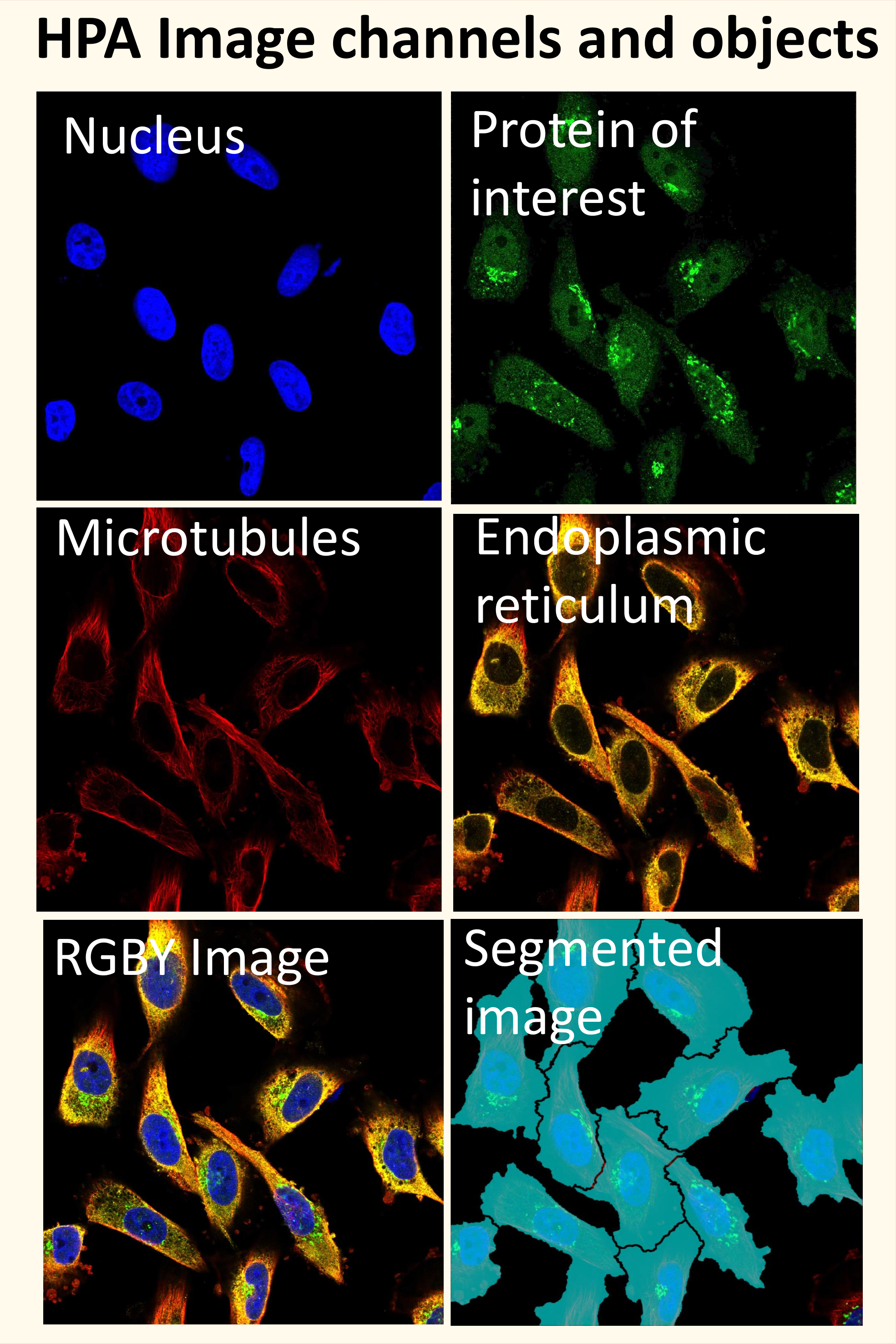}
         \label{fig:fourch}
     \end{subfigure}
      \hfill
    \vspace{-0.3cm}
     \begin{subfigure}[b]{0.375\textwidth}
         \centering
         \caption{}
         \includegraphics[width=\textwidth]{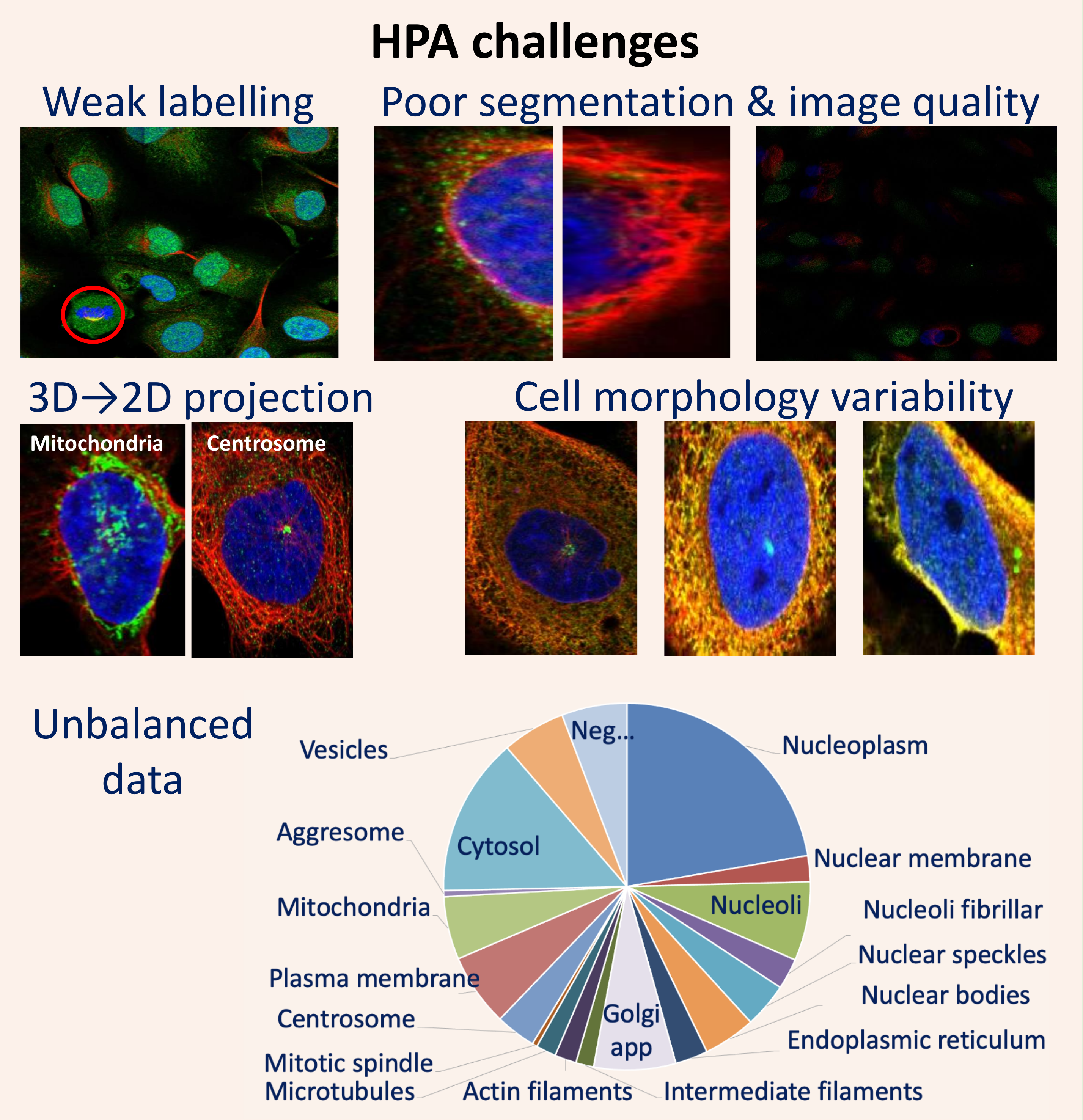}
         \label{fig:challenges}
     \end{subfigure}
      \hfill
      \vspace{-0.4cm}
     \begin{subfigure}[b]{\textwidth}
         \centering
         \caption{}
         \vspace{-0.3cm}
         \includegraphics[width=\textwidth]{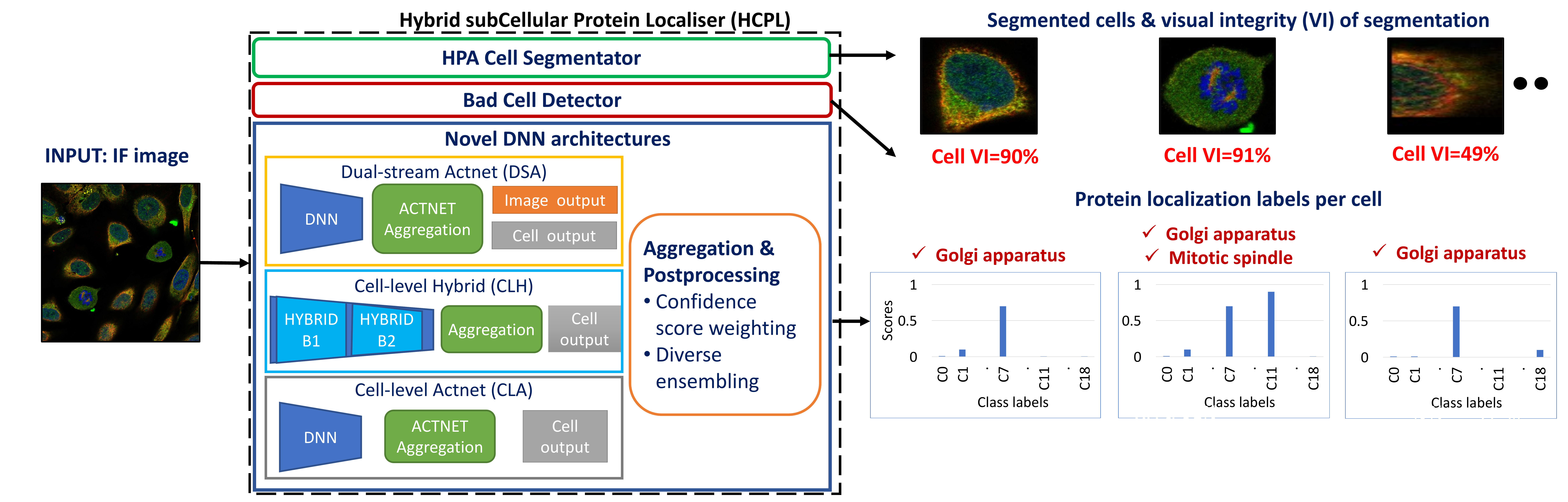}
         \label{fig:dnn1}
     \end{subfigure}
        \hfill
     \begin{subfigure}[b]{0.47\textwidth}
         \centering
        \vspace{-0.7cm}
         \caption{}
        \vspace{-0.2cm}
         \includegraphics[width=\textwidth]{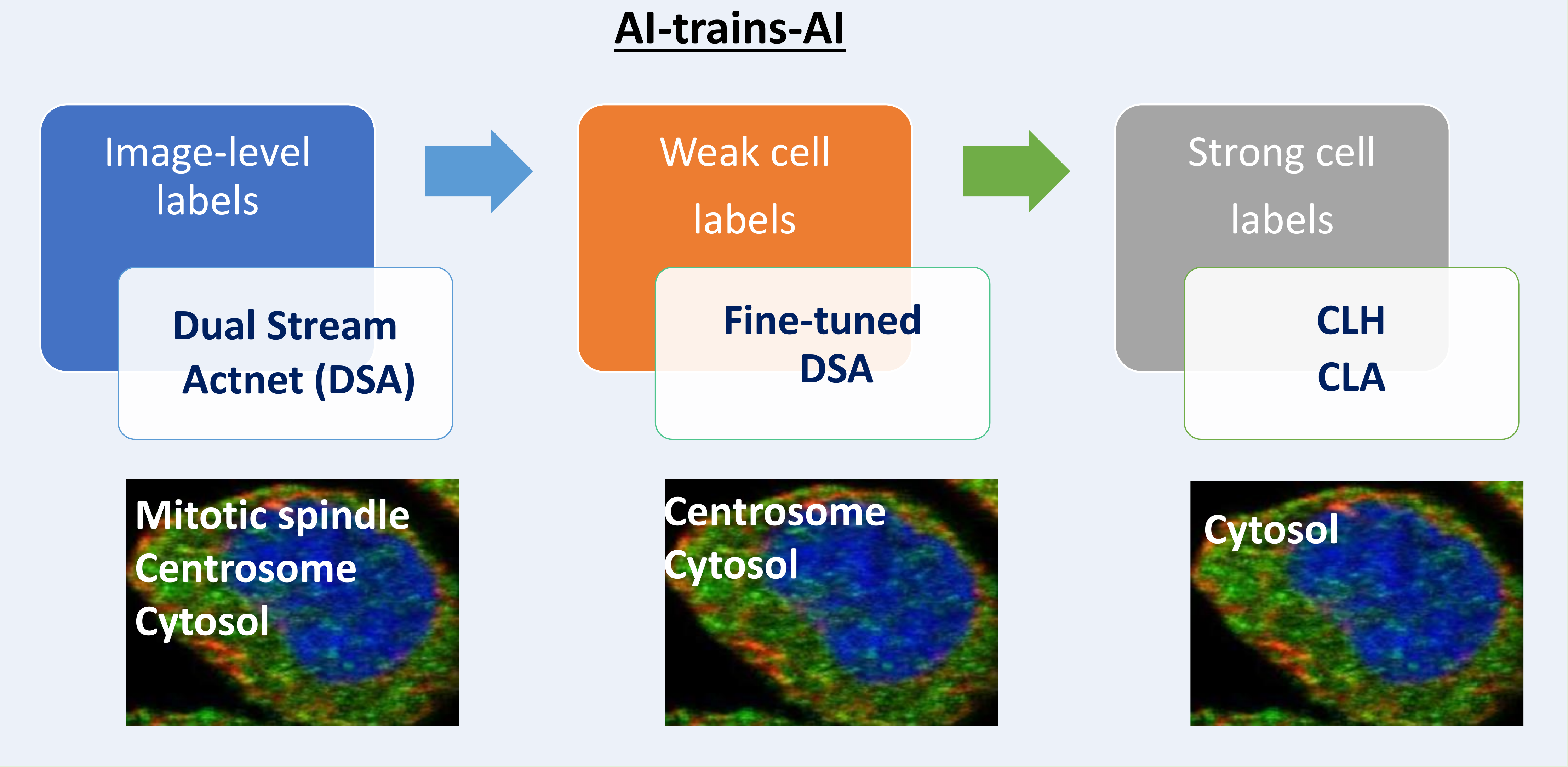}
         \label{fig:dnn3}
     \end{subfigure}
        \hfill
     \begin{subfigure}[b]{0.52\textwidth}
         \centering
        \vspace{-0.7cm}
         \caption{}
              \vspace{-0.2cm}
         \includegraphics[width=\textwidth]{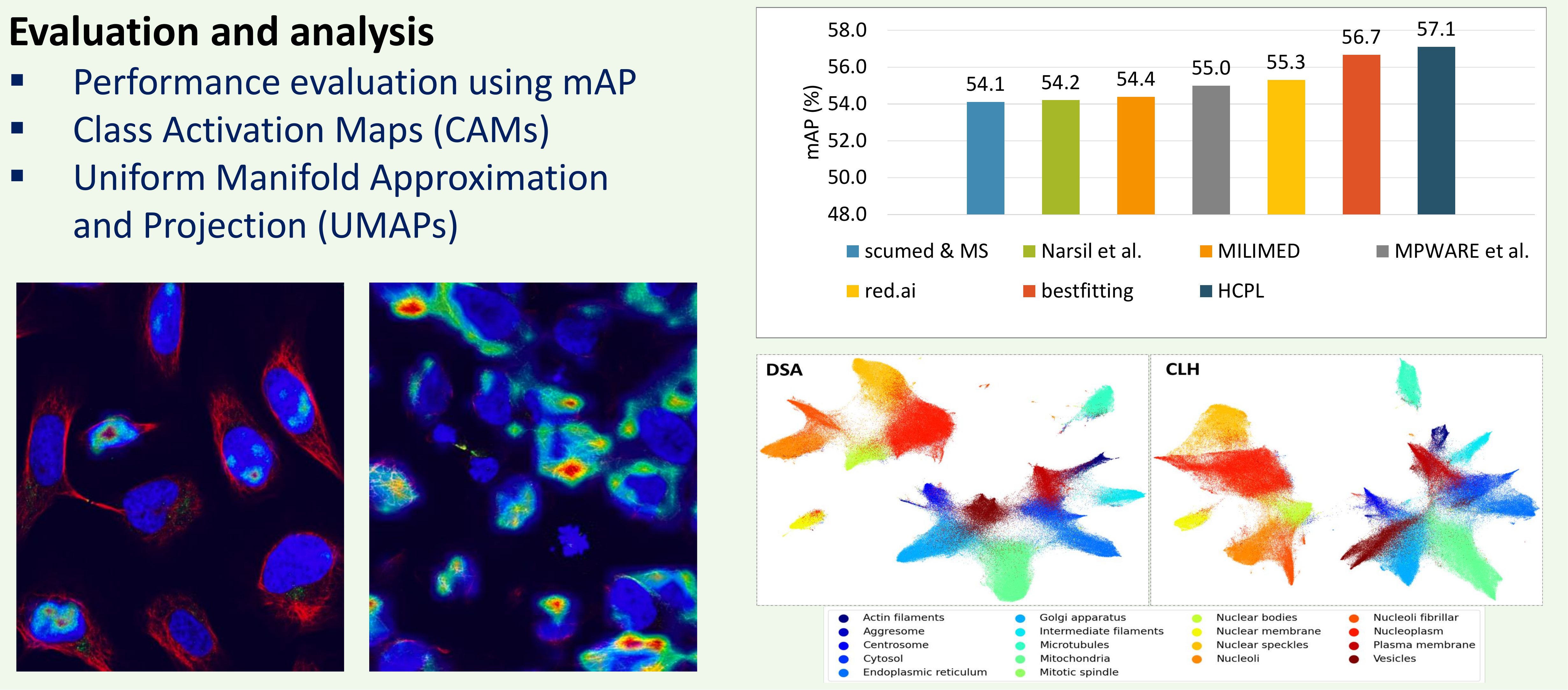}
         \label{fig:dnnana}
     \end{subfigure}
      \vspace{-0.5cm}
        \caption{\textbf{Overview of the HPA challenge and the proposed solution.} \textbf{a}, HPA aims to provide a repository of information on all proteins in the human body, from cells to organs. The location of proteins is examined by immunofluorescence (IF) staining followed by confocal microscopy imaging. \textbf{b}, Each HPA image is represented by four channels, the nucleus (blue), the protein of interest (green), microtubules (red) and the endoplasmic reticulum (yellow). \textbf{c}, The aim is to build ML models to predict protein organelle localisation labels for each cell. The primary challenge is weak labelling: here the image is labelled as mitotic spindle, but only a single cell (circled in red) represents this label. Other difficulties are challenging imaging conditions and the quality of cell segmentation. Furthermore, projections of 3D cell structures onto 2D images often cause ambiguities: here structures lying above but outside the nucleus (mitochondria or centrosome) appear within the nucleus on some images. The dataset also has a severe class imbalance. \textbf{d}, Our HCPL system takes IF images as input and outputs segmented cells, protein localisation labels with associated probabilities and the visual integrity scores for each cell.
        \textbf{e}, The quality of cell labels is improved using the robust AI-trains-AI algorithm. \textbf{f}, Experimental evaluation demonstrates state-of-the-art single-cell classification performance of HCPL. Detailed analysis is performed using CAM and UMAP plots.}
    \label{fig:ARCH}
\end{figure*}

To address this gap, we have developed a novel deep-learning-based system, the Hybrid subCellular Protein Localiser (HCPL), for robust protein localisation with subcellular resolution. Our method is the first to characterise single-cell heterogeneity in extensive collections of microscope images by producing precise annotations of subcellular protein localisation patterns in individual cells. Our system uses an ensemble of diverse deep architectures embedded within a dual-head structure that learns effectively from images of either multiple cells annotated at an image level or individually annotated cells. We develop and validate our approach using the Human Protein Atlas (HPA) \cite{hpadataset}, which is the largest public dataset and forms an invaluable resource for studying cell biology (Methods). Importantly, the HPA contains a large collection of images depicting specific protein localisations at a subcellular level, acquired using immunofluorescence staining followed by confocal microscopy imaging \cite{subcelular} (Figure \ref{fig:confocal}). The images use a four-channel format as described in Figure \ref{fig:fourch}.
This resource is key for understanding human cells, and the complex molecular mechanisms underpinning their functions \cite{Emma, Christopher}, taking advantage of antibody-based multiplexed protein imaging methods \cite{bodenmiller2016multiplexed, Hickey}. 

Aimed at individual cell analysis, the HCPL system successfully addresses several major challenges (Figure \ref{fig:challenges}). Compared to currently available methods that can only provide predictions for a collection of cells (i.e. at the image-level), the requirement to classify each cell individually is a far more difficult task. While single-cell localisation requires accurate predictions for each cell, on an image level, it is sufficient to locate one relevant cell where the presence of a protein is most evident without classifying each of the remaining cells individually. From a ML perspective, this difficulty is further compounded by the frequent lack of accurate ground truth for training with typically only image-level labels available. For example, each HPA image comprises many cells jointly labelled with one set of labels, defined as the union of individual cell labels. Hence, the image-level labels are incorrect for some of the cells in an image; this phenomenon is called weak labelling. Further difficulties arise from a dramatic variability in cells' morphology, as well as from the use of different cell lines and inconsistent cell image quality caused by staining or segmentation failures. Finally, we must also contend with extremely imbalanced frequencies of the localisation classes, along with the multi-label setting where a single cell can take multiple labels.

Importantly, the best-published approaches were found unable to handle these challenges. We show that image-level algorithms \cite{kagglenature} perform poorly on the task of cell-level protein localisation, achieving circa 33\% mAP (mean Average Precision) (Methods). A recently presented unsupervised approach \cite{cytoself} can learn latent space representations that loosely correspond to specific protein localisation patterns; however, this method cannot be used to \textit{predict} such patterns. The proposed HCPL, tested on the gold-standard HPA dataset \cite{hpadataset}, achieves a performance of 57.1\% mAP, which defines the state-of-the-art for single-cell classification.

To better understand our system's operation and the benefits brought by its novel components, we perform extensive testing, including a series of ablation studies. Furthermore, we benchmark our system against the leading solutions developed during the recent Kaggle competition (Human Protein Atlas - Single Cell Classification \cite{HPA}). HCPL is superior to the best results achieved in that competition. Finally, the work is concluded with an analysis that verifies the biological correctness and meaningfulness of the systems' predictions.

We believe the HCPL system fills an important gap and is well-placed to contribute to our knowledge of spatial biology in health and disease, and its application to the development of therapeutics.

\section*{Results} \label{results}

\noindent\newline\textbf{HCPL - Hybrid subCellular Protein Localiser.} 
An overview of the typical lab workflow, the HPA challenge and our HCPL solution are presented in Figure \ref{fig:ARCH}. 
The HCPL system receives multi-channel images (Figure \ref{fig:fourch}), segments individual cells using the HPA Cell Segmentator (Methods), and analyses each cell in turn to estimate both its visual integrity and the probabilities of proteins being present in specific subcellular compartments. HCPL combines several novel DNNs (Deep Neural Networks) to capture the biological variability and richness of patterns present in the HPA data, specifically:
\begin{itemize}
\item A multi-task Dual-stream Actnet (DSA) DNN, which learns to extract and fuse information from both images and individual cells. The DSA mitigates the vast amounts of false-positive cell predictions caused by the weak cell labelling used for training. 
\item A robust Cell-level Hybrid model (CLH), which combines learnt deep features with handcrafted features from a set of well-defined filters.
\item A highly performing Cell-level Actnet (CLA), which employs learnable parametric activations to robustly aggregate discriminatory image features.
\end{itemize}

We develop an innovative method to select and optimally ensemble multiple classifiers to fully benefit from the diversity in expert opinions provided by different DNNs. Our HCPL system ensembles nine diverse architectures and achieves protein localisation accuracy of 57.1\% mAP, which defines the state-of-the-art. Execution time is below 1 second per cell image on a single thread CPU and NVIDIA-P100 GPU. 

For system training, we propose an ''AI-trains-AI'' framework to improve the quality of the weak labels (Figure \ref{fig:dnn3}). It employs two techniques: (1) DNN-powered re-labelling, where labels are iteratively improved during the training, and 
(2) automatic adjustment of localisation confidence factors based on a cell's estimated visual integrity, limiting the impact of cells with segmentation errors.

\begin{figure*}
     \centering
     \begin{subfigure}[b]{1\textwidth}
         \centering
         \caption{}
         \includegraphics[width=\textwidth]{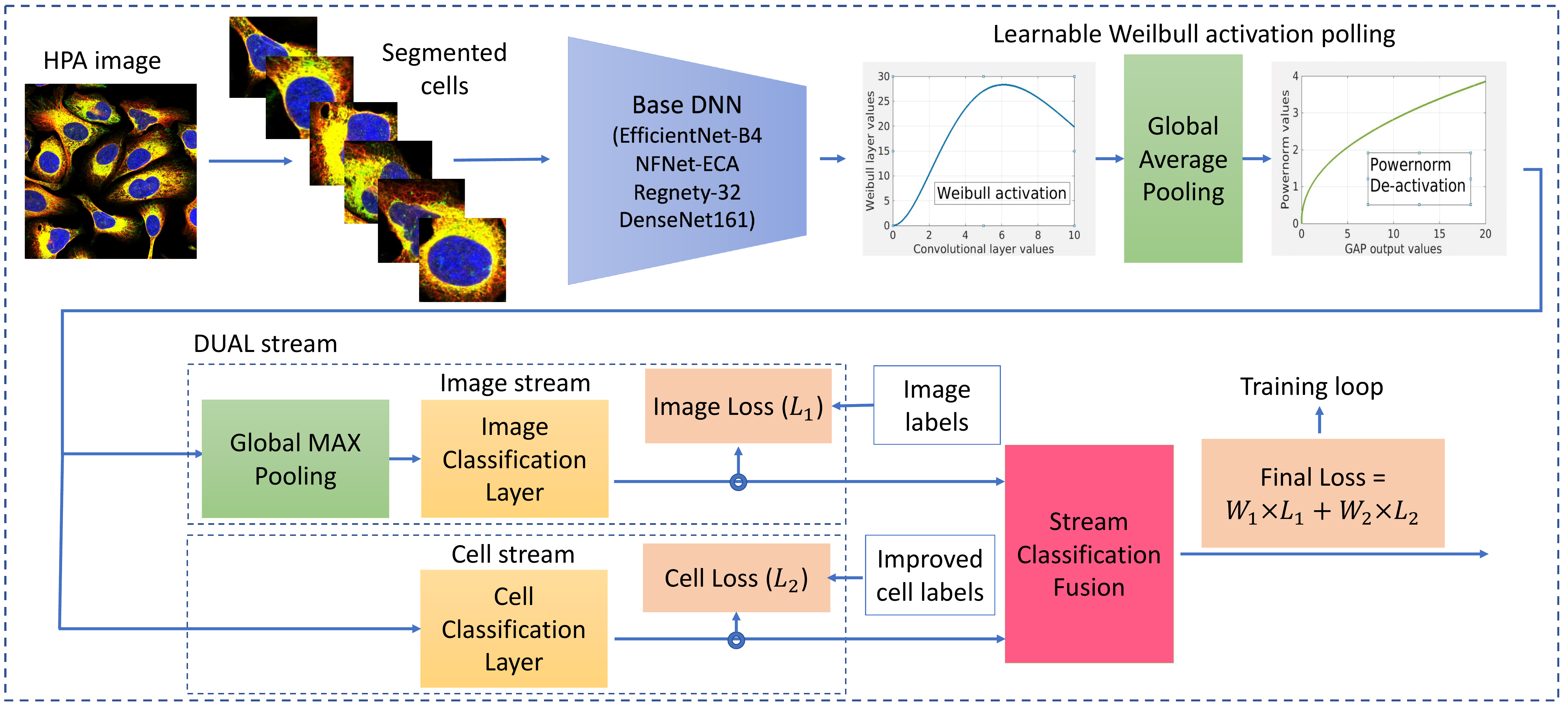}
        \vspace{-0.6cm}
         \label{fig:dualnet}
     \end{subfigure}
     \begin{subfigure}[b]{1\textwidth}
         \centering
         \caption{}
         \includegraphics[width=\textwidth]{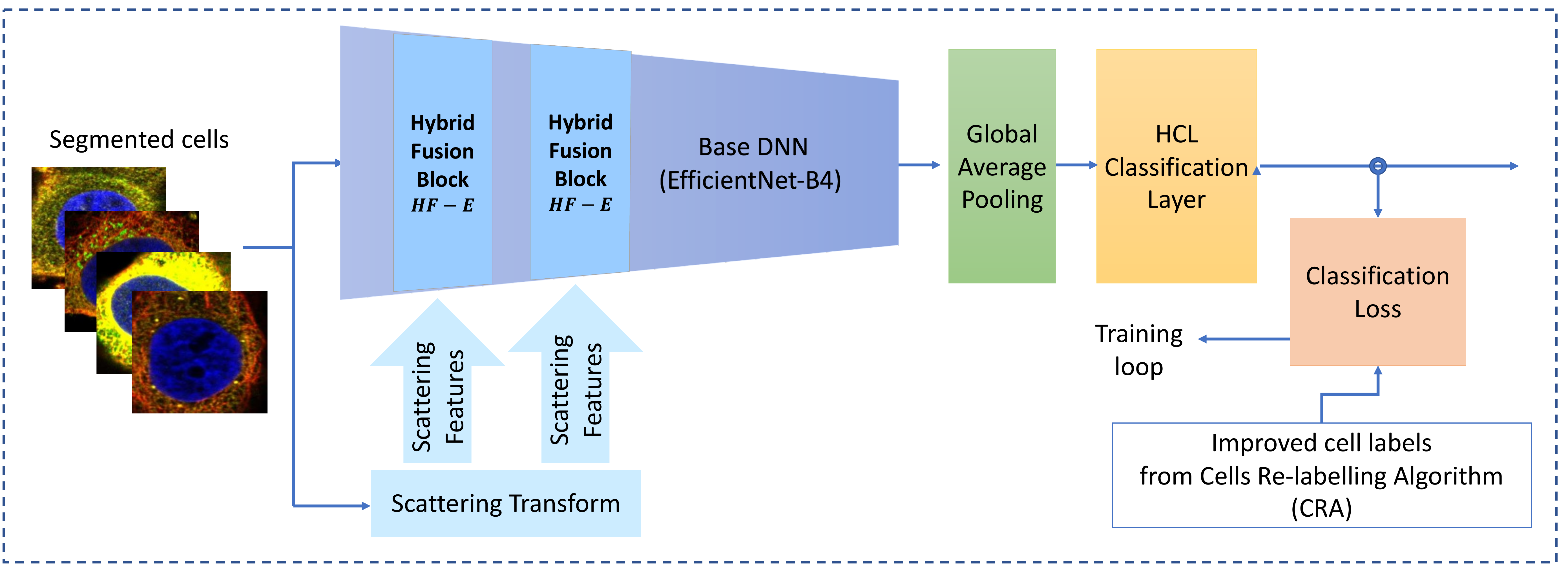}
            \vspace{-0.6cm}
        \label{fig:hybrid}
     \end{subfigure}
     \begin{subfigure}[b]{1\textwidth}
         \centering
        \caption{}
         \includegraphics[width=\textwidth]{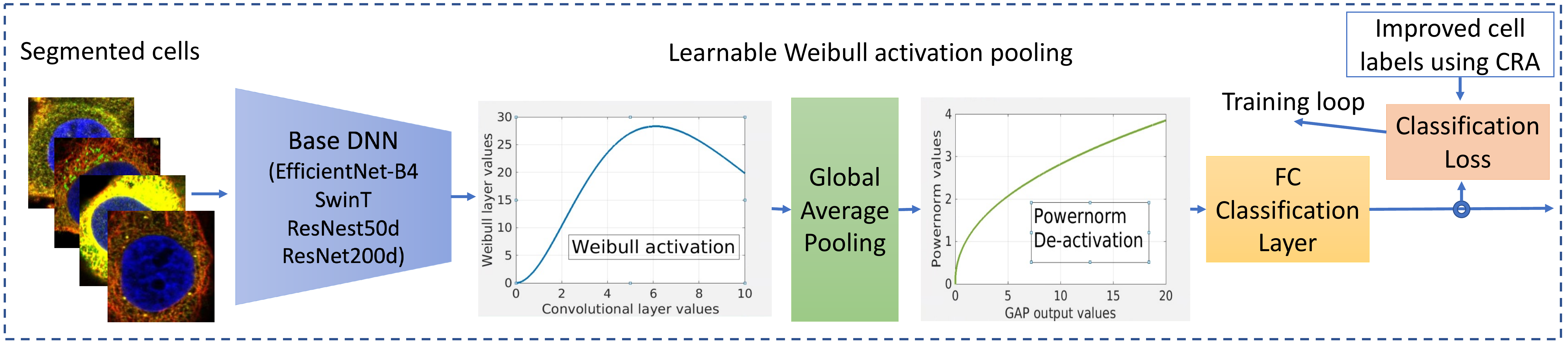}
         \label{fig:actnet}
     \end{subfigure}
        \caption{\textbf{Key components of the HCPL system}. \textbf{a}, The DSA comprises an image stream that models the entire population of cells and a cell stream that classifies patterns in each cell in the image. \textbf{b}, CLH is a single stream inductive hybrid architecture that uses a scattering transform as a complimentary source of information about each cell. \textbf{c}, CLA is a single stream architecture which employs Weibull activation pooling to aggregate deep features before classification.}
\end{figure*}

\noindent\newline\textbf{DSA model for efficient cell-level and image-level information fusion.}
To maximise classification performance and increase robustness, our DSA model (Figure \ref{fig:dualnet}) jointly exploits local (cell-level) and global (image-level) visual cues. The DSA architecture comprises a baseline DNN (e.g. EfficientNet \cite{pmlr}) followed by the dual-stream network. From each image, cells are extracted and forwarded to the DNN component, producing deep convolutional features. The features are passed to a learnable Weibull activation Pooling \cite{WAP} to improve the discriminative power of the feature map (Methods). Specifically, the weak uninformative features are dampened, and a learnable proportion of strong informative features are equalised. The aggregated dense features are classified in two parallel streams - the image stream outputs predictions for the entire population of cells in an image, whereas the cell stream generates a set of predictions for each cell.

The DSA network is trained end-to-end using a weighted sum of Binary Cross-Entropy (Methods) losses from cell and image streams. At inference, for each cell, the probability of each class is computed as a product of relevant image and cell stream probabilities. For selected classes where image-level predictions are less reliable, the final probabilities given for such classes are the cell stream probabilities.

\begin{figure*}
     \centering
     \begin{subfigure}[b]{0.9\textwidth}
         \centering
          \caption{}
         \includegraphics[width=\textwidth]{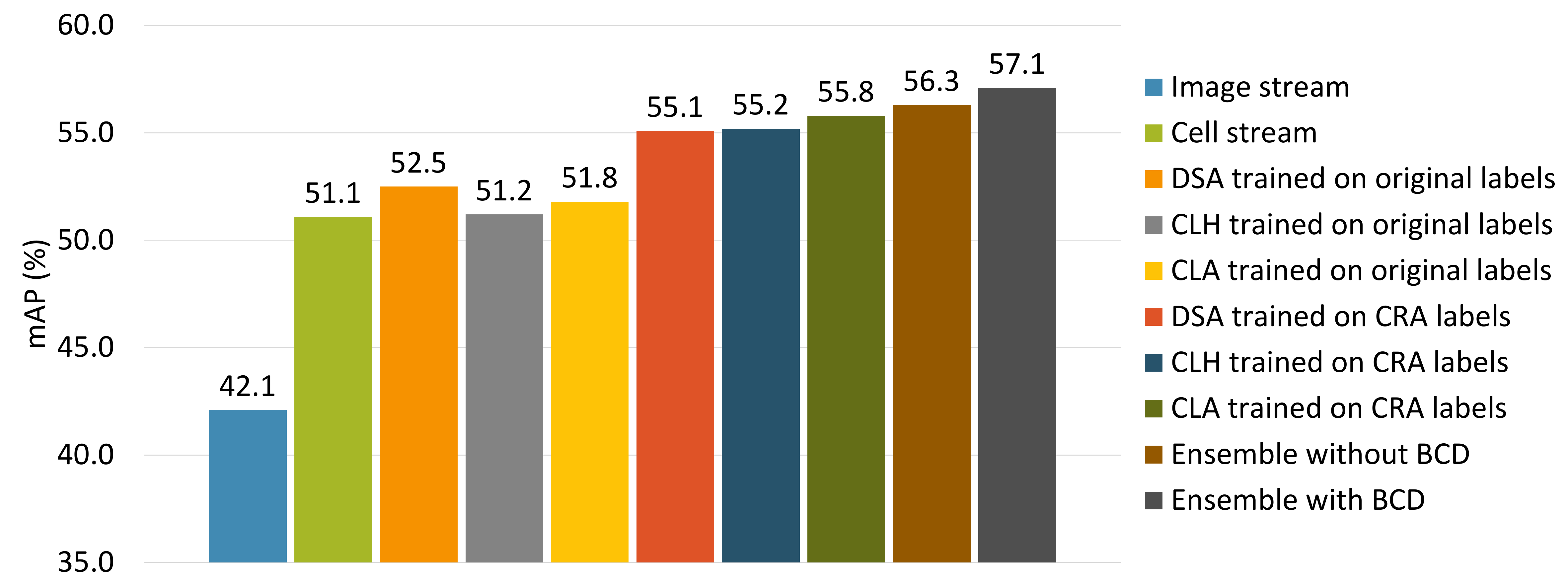}
                 \vspace{-0.1cm}
         \label{fig:stream}
     \end{subfigure}
     \hfill
        \begin{subfigure}[b]{\textwidth}
         \centering
         \caption{}
         \includegraphics[width=\textwidth]{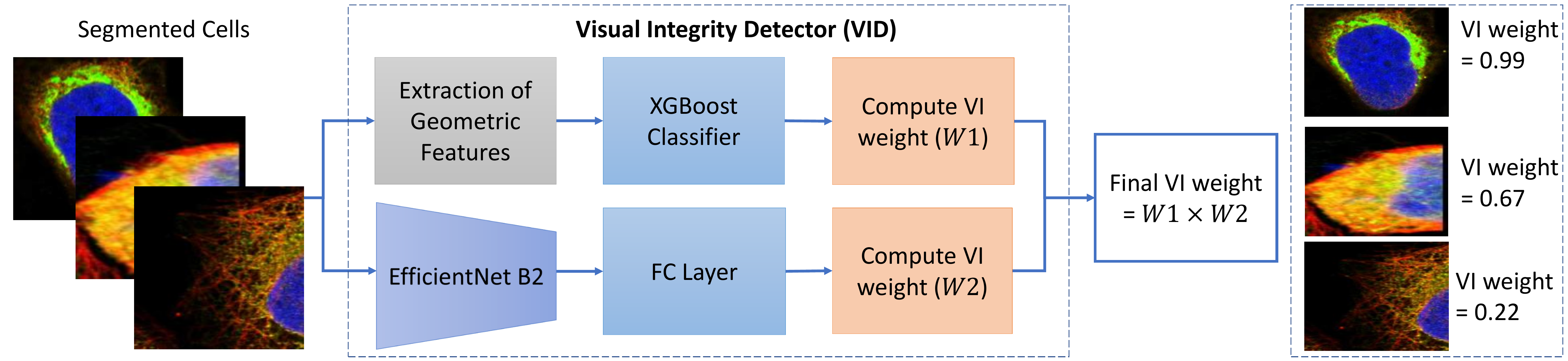}
         \label{fig:bcp}
     \end{subfigure}
     \begin{subfigure}[b]{\textwidth}
         \centering
          \caption{}
         \includegraphics[width=0.95\textwidth]{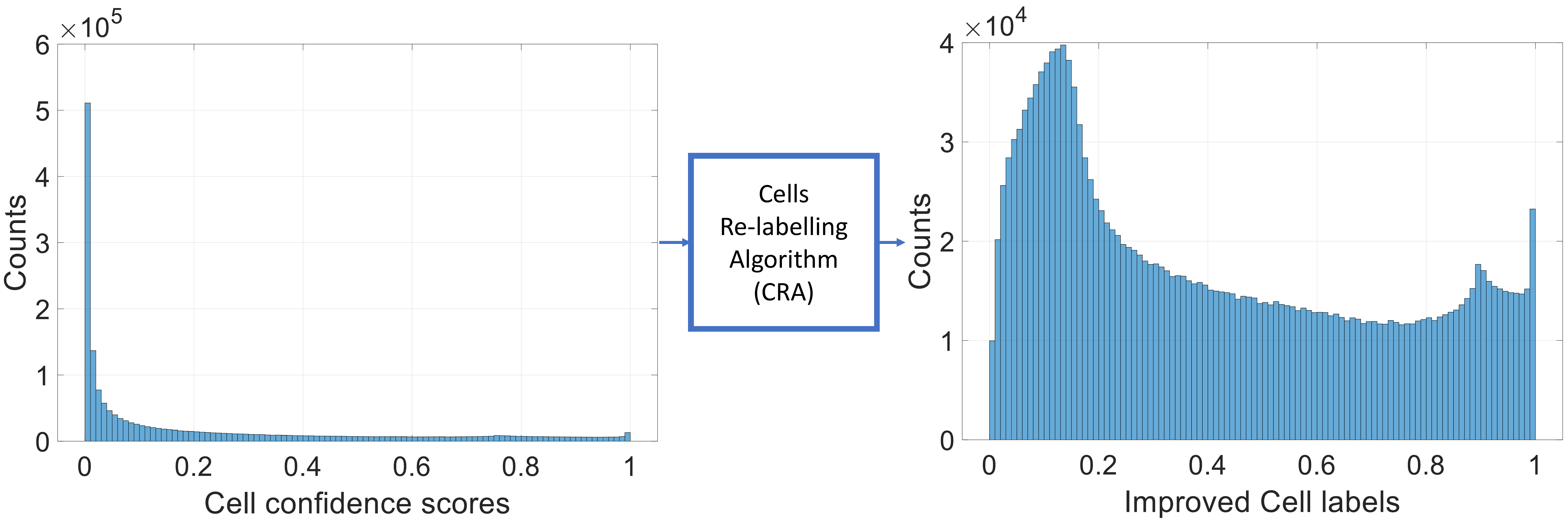}
         \label{fig:histogram}
     \end{subfigure}
        \caption{\textbf{Performance and functional details of selected elements of our system}. \textbf{a}, The classification performance of different modules of HCPL including image stream, cell stream, DSA, CLH, CLA, Ensemble without VID system and Ensemble with VID system. \textbf{b}, The VID system consists of two modules: one that extracts geometric features from a cell and employs an XGBoost classifier, and a DNN to predict the visual integrity of the cell. \textbf{c}, The figure on the left shows the histogram of confidence scores assigned to labels based on the ground-truth data. Approximately 60\% of labels have scores below 0.1 indicating two issues: (1) DNNs are trained using weak labels, (2) training of DNNs is complex due to segmentation faults and cell variability. The CRA transforms the scores using a power-normalisation operation to generate a new set of probabilities $K$. The probabilities $K$ then replace their associated labels, producing a new set of improved continuous valued cell labels. The histogram of improved cell labels is presented on the right. }
        \label{fig:three graphs}
\end{figure*}

Figure \ref{fig:stream} shows that the image stream individually achieves 42.1\% mAP. This baseline performance comes from mapping image-level labels to all cells in that image, resulting in vast over-labelling. The cell stream achieves a better 51.1\% mAP, still relatively low due to the weak labels used in training. However, our dual-stream DSA architecture achieves 55.1\%, a gain of (+4\%) stemming from the intelligent fusion of both streams.

\noindent\newline\textbf{Cell-level Hybrid and Actnet models for capturing cell variability.}
We develop two innovative and high-performing cell-level architectures to comprehend the richness of patterns in the HPA. The first network, CLH, uses an inductive hybrid system (Figure \ref{fig:hybrid}) to fuse hand-crafted features extracted by a scattering transform with features learnt from the data. The scattering transform, formed by cascading wavelets, guarantees translation invariance and a linear response to deformations. The complementary nature of mathematically well-defined and data-adaptable filters yield a robust model. To ensure these properties are propagated into the DNN flow, two Hybrid Fusion blocks are inductively integrated into an EfficientNet architecture \cite{Dmitry}. The Hybrid-DNN module works on cell-level images and outputs convolutional features, which are aggregated and passed to the classification module to generate predictions.

The second network, CLA, uses parametric activations for adaptive and robust aggregation (Figure \ref{fig:actnet}). At its core, a base DNN extracts features which are fed to a learnable activation layer \cite{WAP} comprising the Weibull function. The Weibull function amplifies responses corresponding to distinctive features of cells that are important for the classification tasks relative to the background. The transformed feature vectors are forwarded to a global average pooling layer, power-normalisation layer and classification module to compute predictions.

CLH and CLA are trained using a weighted Binary Cross-Entropy loss, where rare classes are given more weight than frequent classes. At inference, the class probabilities for each cell are computed similarly to the DSA, i.e. as a product of relevant DSA image stream and cell-level stream probabilities (CLA or CLH outputs), conditioned on the reliability of the image stream.

A class-wise comparison demonstrates that the CLH better represents rare classes (0.22\% and 0.11\% mAP gain over CLA on aggresome and mitotic spindle). The main gain comes from the handcrafted wavelet filters at the base of the Hybrid Fusion Block, which help the network generalise with fewer training examples. Conversely, the CLA's ability to select the most discriminative features helped it perform better on the visually similar intermediate filaments (+0.13\%) and actin filaments (+0.31\%). We observe in Figure \ref{fig:stream} that cell-level architectures achieve better mAP than the DSA.

\noindent\newline\textbf{Estimation of cell visual integrity with VID.} 
Poor imaging or segmentation failures lead to errors in classification. Hence, an important feature of our system is a Visual Integrity Detector (VID) that is trained to detect cell capture errors and to adjust classification confidence accordingly. It is generally considered that for a meaningful prediction, at least 50\% of a cell has to be captured \cite{HPA}. 

As presented in Figure \ref{fig:bcp}, VID consists of two modules: (i) the first extracts eight morphological features from each cell and uses an XGBoost classifier \cite{XGboost} to evaluate its structural integrity. (ii) An EfficientNet-B2 \cite{pmlr} network, which predicts the ratio of the total cell body being captured. We developed a cell dataset for training the VID, which is made available for the community (Methods).

The results in Figure \ref{fig:stream} show that the inclusion of the VID module improves the system performance by 0.8\% mAP.

\noindent\newline\textbf{Improving label quality using an ``AI-trains-AI'' approach.} \label{remining}
We leverage the generalisation ability of our DSA model to learn from noisy data and assign confidence scores to ground-truth labels (image-level labels naively mapped to cells). A confidence score represents the probability that a ground-truth label is correct; hence it takes values in $[0, 1]$. We use confidence scores to perform iterative training where subsequent models focus less on low-confidence cell labels while emphasising those with high confidence. This is achieved by our Cells Re-labelling Algorithm (CRA). It first computes the per-cell component probabilities using three DSAs trained on weak labels (bases EfficientNet-B4, NFNet-ECA \cite{nfnet} and ResNest50d \cite{ResNeSt}). These three component probabilities are averaged across each label for each cell to obtain a new set of combined confidence factors.

In the second step, we re-evaluate the true-positive labels for each cell based on these combined confidence factors (Figure \ref{fig:histogram} left). The CRA transforms the factors using a power-normalisation operation ($c \mapsto c^\beta$, $c$ is a probability) to generate a new set of probabilities which replace the original labels, yielding improved continuous-valued cell labels (Figure \ref{fig:histogram} right). The CRA effectively re-evaluates the weak cell labels originally inherited from the image-level labels. Next, we retrain the DSA model using re-labelled cells.

Results in Figure \ref{fig:stream} show that the DSA retrained on improved labels achieved performance of 55.2\% mAP compared to DSA trained on weak labels (52.5\% mAP). The above process is employed twice to refine label quality. Finally, the cell labels obtained after round two are used to train the CLA and CLH models, leading to improvements exceeding +4\% mAP over the models trained on the original labels (Figure \ref{fig:stream}). 

\begin{figure*}
     \centering
     \begin{subfigure}[b]{0.95\textwidth}
         \centering
        \vspace{-0.4cm}
         \caption{}
         \vspace{-0.2cm}
         \includegraphics[width=\textwidth]{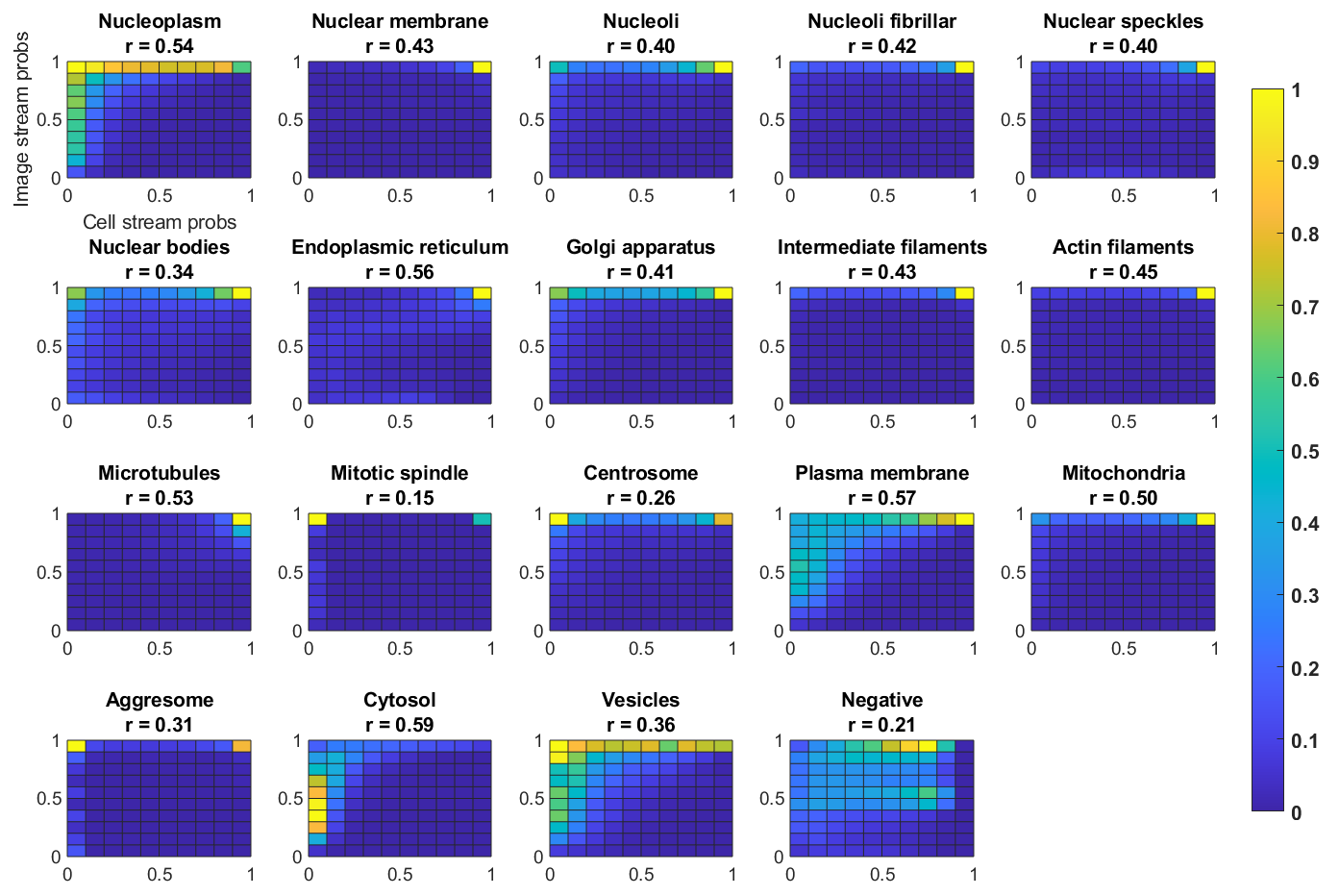}
        \vspace{-0.8cm}
         \label{fig:img_cell_corr}
     \end{subfigure}
          \hfill
     \begin{subfigure}[b]{0.49\textwidth}
         \centering
         \caption{}
         \includegraphics[width=\textwidth]{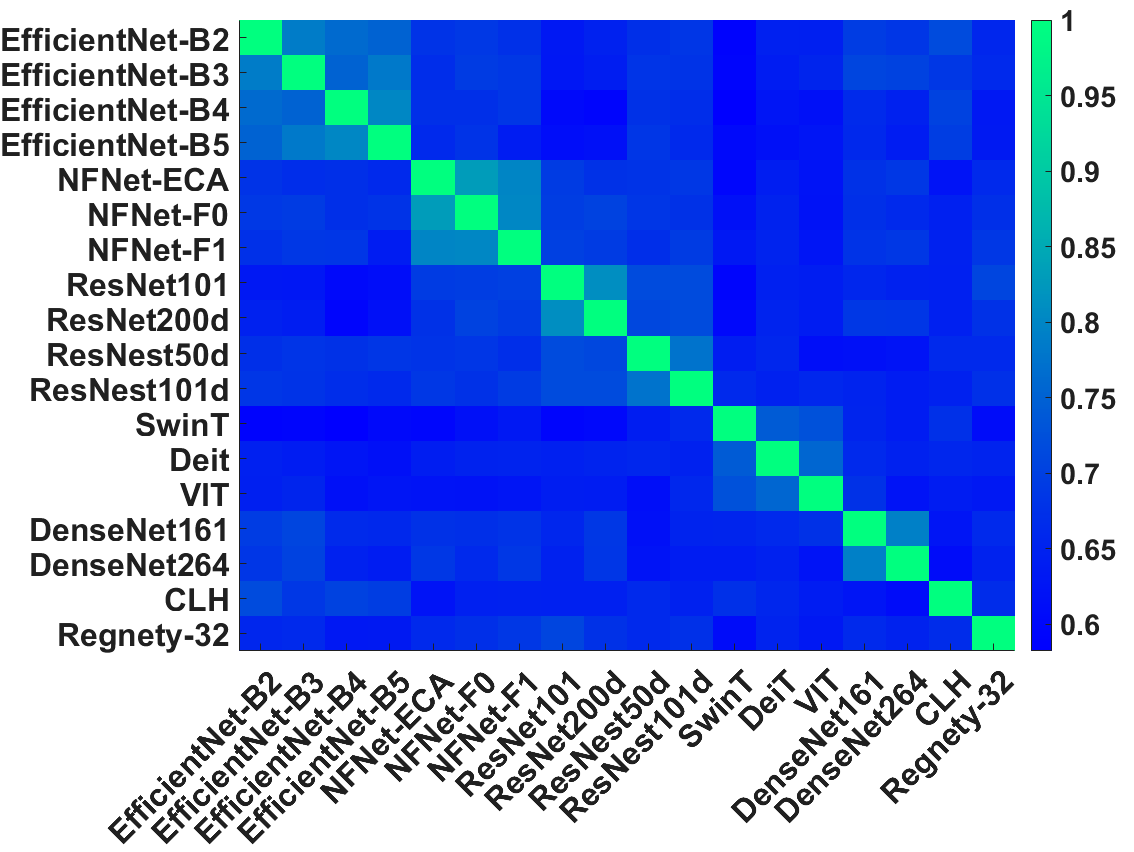}
         \label{fig:model_corr}
    \end{subfigure}
        \hfill
     \begin{subfigure}[b]{0.49\textwidth}
         \centering
         \caption{}
         \includegraphics[width=\textwidth]{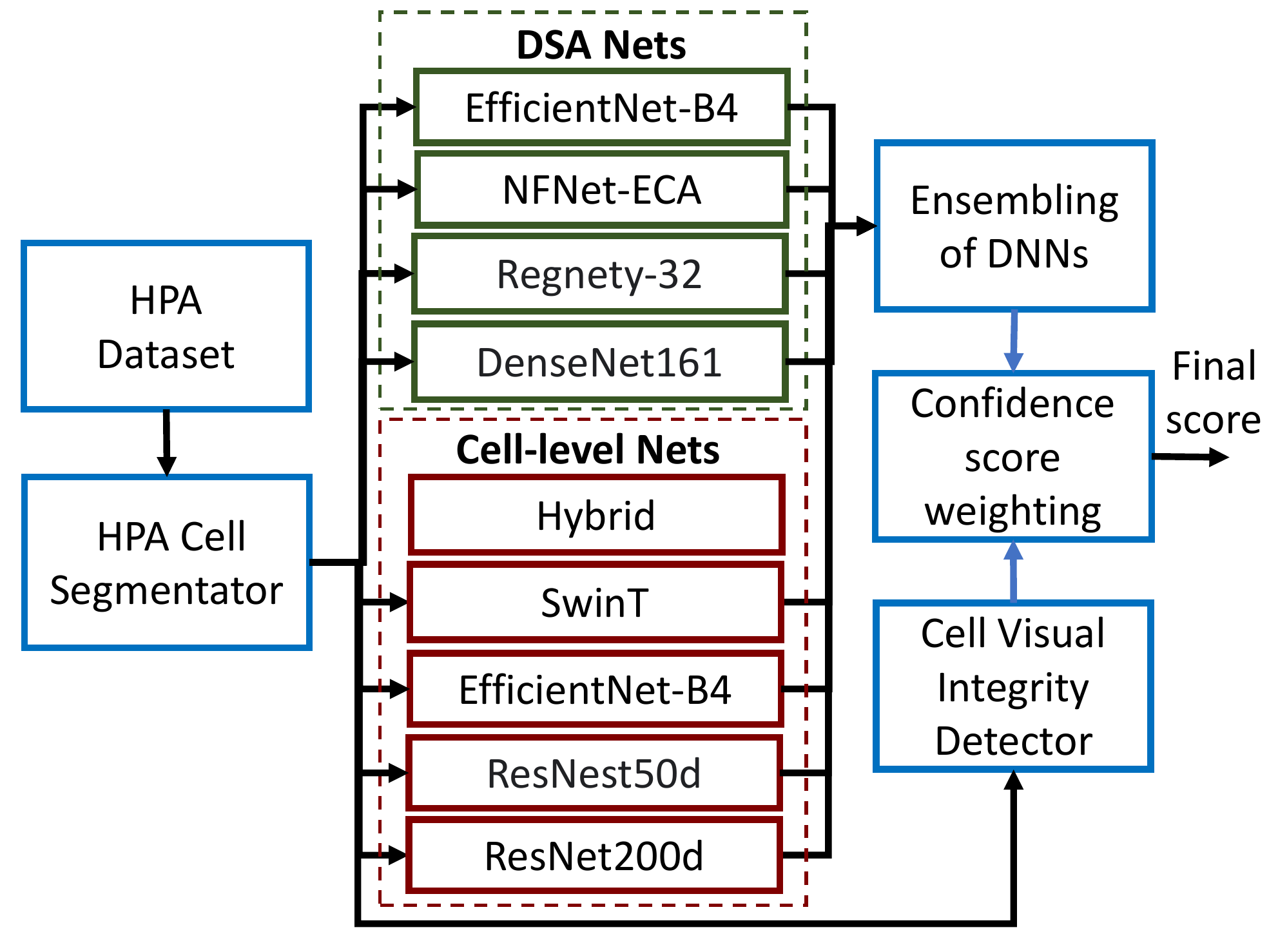}
         \label{fig:inference}
     \end{subfigure}
             \vspace{-0.3cm}
        \caption{\textbf{Correlation based ensembling of diverse DNNs}. \textbf{a}, 2D histograms of image-level and cell-level predictions for all classes. Classes mitotic spindle, centrosome and aggresome have very low correlation coefficients $r$. The values of $r$ are displayed with the class names at the top of correlation plot. The image stream labels the majority of cells as having aggresome, visible as a bright spots at the first row. However, we know that only around 30\%-60\% of cells in an image have aggresome highlighted on the green channel. The cell-level stream is able to reject cells without aggresome highlighted, causing a bright spot at the top-left corner of the histogram to appear. The same phenomenon can be seen for the class mitotic spindle and centrosome. \textbf{b}, Correlation between confidence scores generated by different DNNs. \textbf{c}, HCPL inference framework employed to generate submissions for Kaggle.}
        \label{fig:three graphs}
\end{figure*}

\noindent\newline\textbf{Multi-stream information fusion.}
We exploit the fusion of information extracted by selected classifiers working on image and cell levels to maximise the system's performance. Figure \ref{fig:img_cell_corr} shows 2D histograms of image-level and cell-level predictions for all nineteen classes. Each histogram is computed using images with their corresponding label.

We note that certain classes (mitotic spindle, centrosome and aggresome) show disagreement in predictions, where a bright spot is present at the top-left corner. This shows cell-level labelling rejecting cells classified as positive by the image-level labelling. 
This phenomenon can be summarised by computing the correlation coefficient $r$ between the image-level and cell-level predictions. The aforementioned classes have very low correlation coefficient values, reflecting that these rare classes are prone to over-labelling by image classifiers.  

Image and cell level fusion is performed as follows. For classes where the correlation is greater than a certain threshold $\rho_{th}$, the localisation probabilities for a cell are obtained as per class products of image-level and cell-level predictions. For classes with low correlation, we assign the cell-level predictions to the final class probabilities. Please note that all cell-level networks are trained using improved cell-level labels.

\begin{figure*}
     \centering
     \vspace{-0.5cm}
     \begin{subfigure}[b]{0.49\textwidth}
         \centering
        \caption{}
         \includegraphics[width=\textwidth]{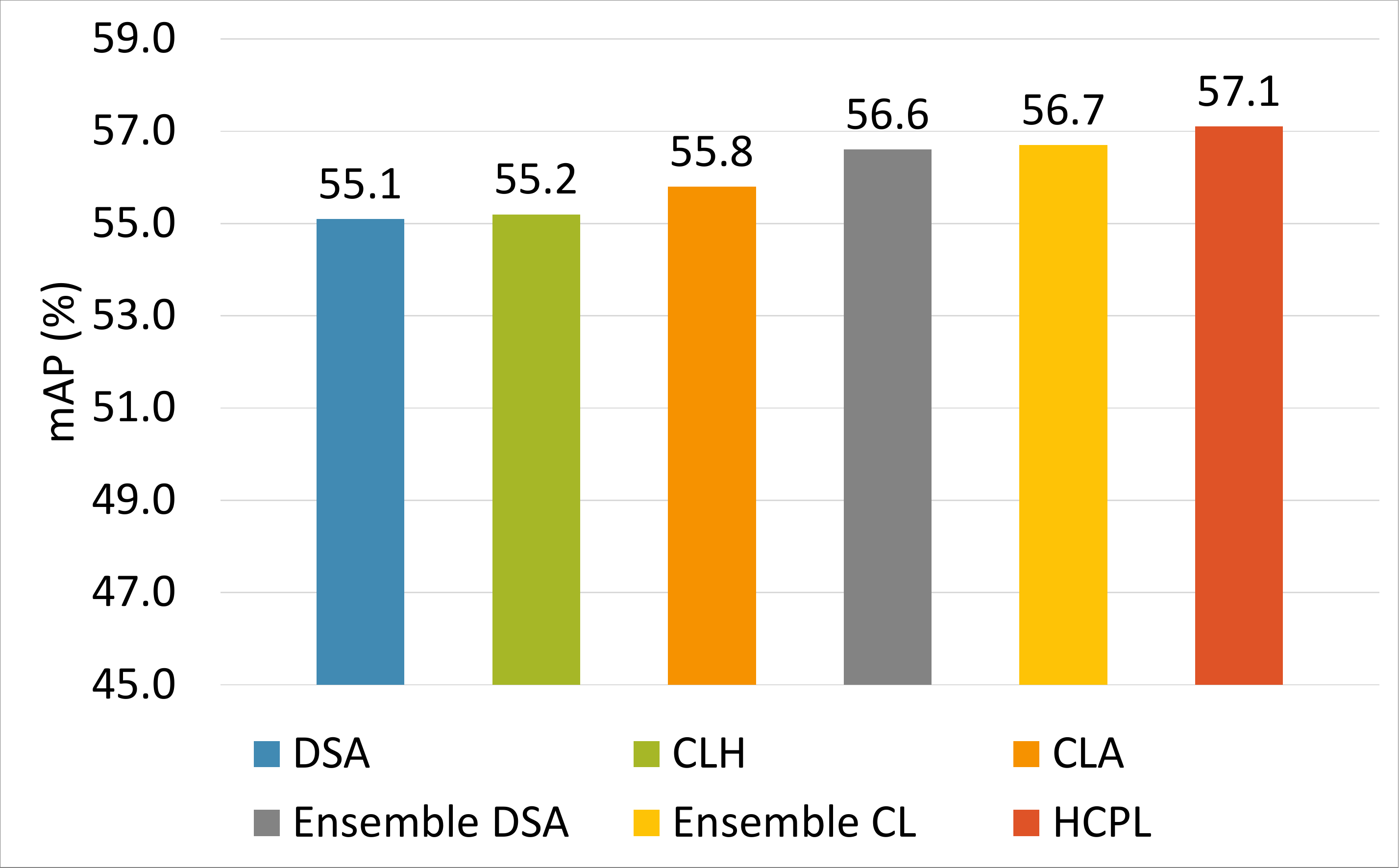}
         \vspace{-0.5cm}
         \label{fig:ens}
     \end{subfigure}
     \hfill
     \begin{subfigure}[b]{0.49\textwidth}
         \centering
        \caption{}
         \includegraphics[width=\textwidth]{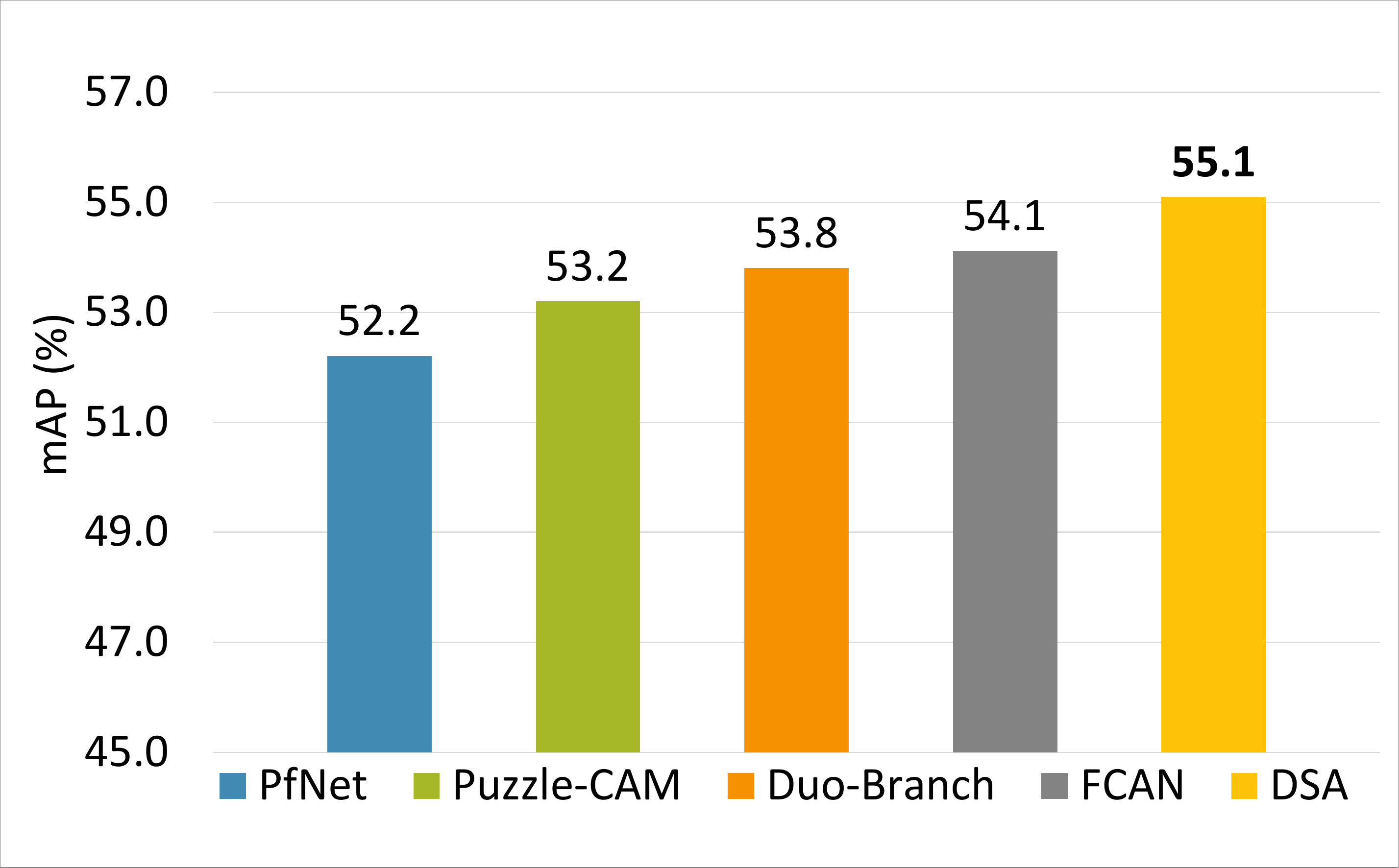}
        \vspace{-0.5cm}
         \label{fig:dual-sota}
     \end{subfigure}
      \hfill
     \begin{subfigure}[b]{0.49\textwidth}
         \centering
         \caption{}
         \includegraphics[width=\textwidth]{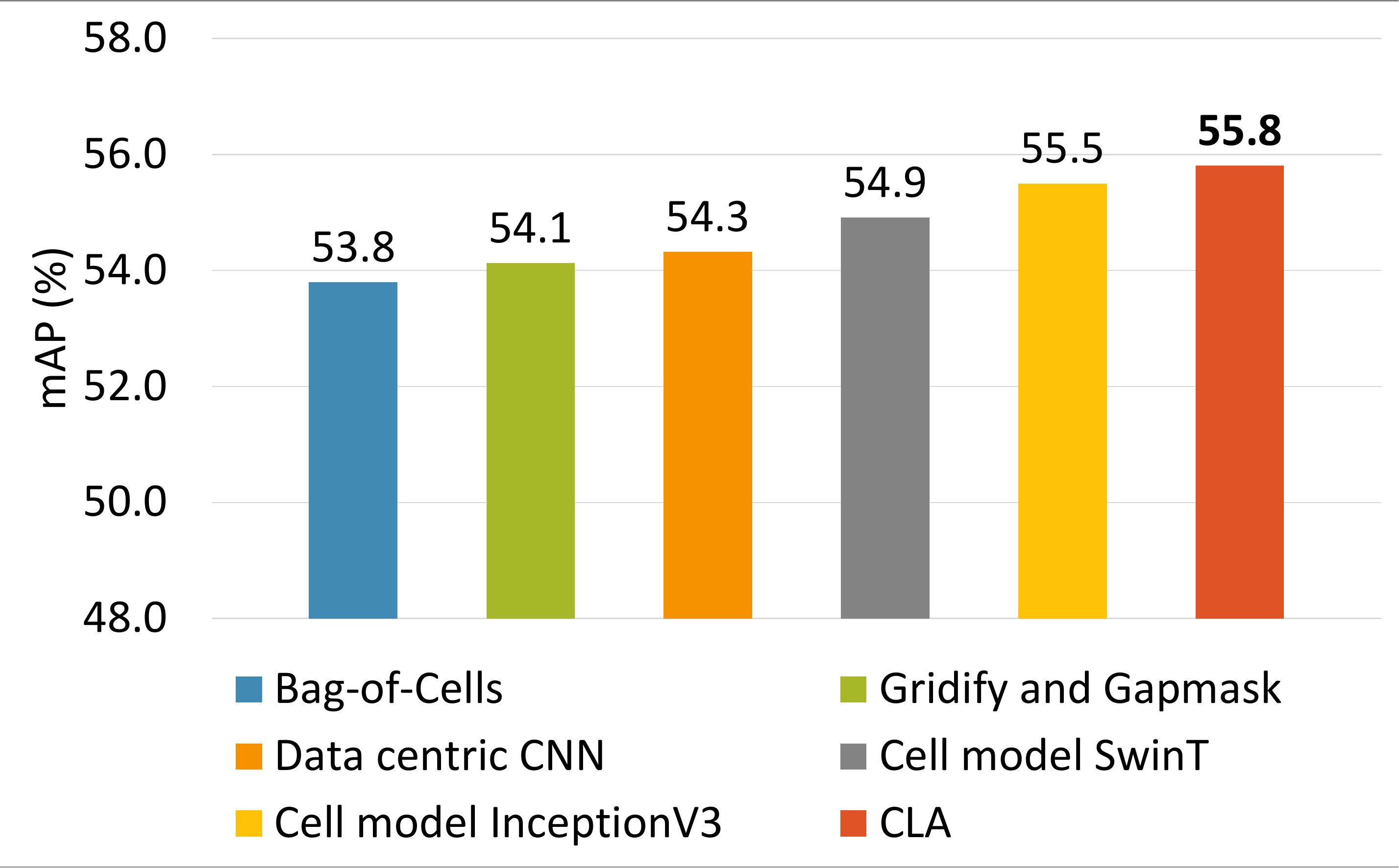}
                 \vspace{-0.5cm}
         \label{fig:cell-sota}
     \end{subfigure}
      \hfill
     \begin{subfigure}[b]{0.49\textwidth}
         \centering
         \caption{}
         \includegraphics[width=\textwidth]{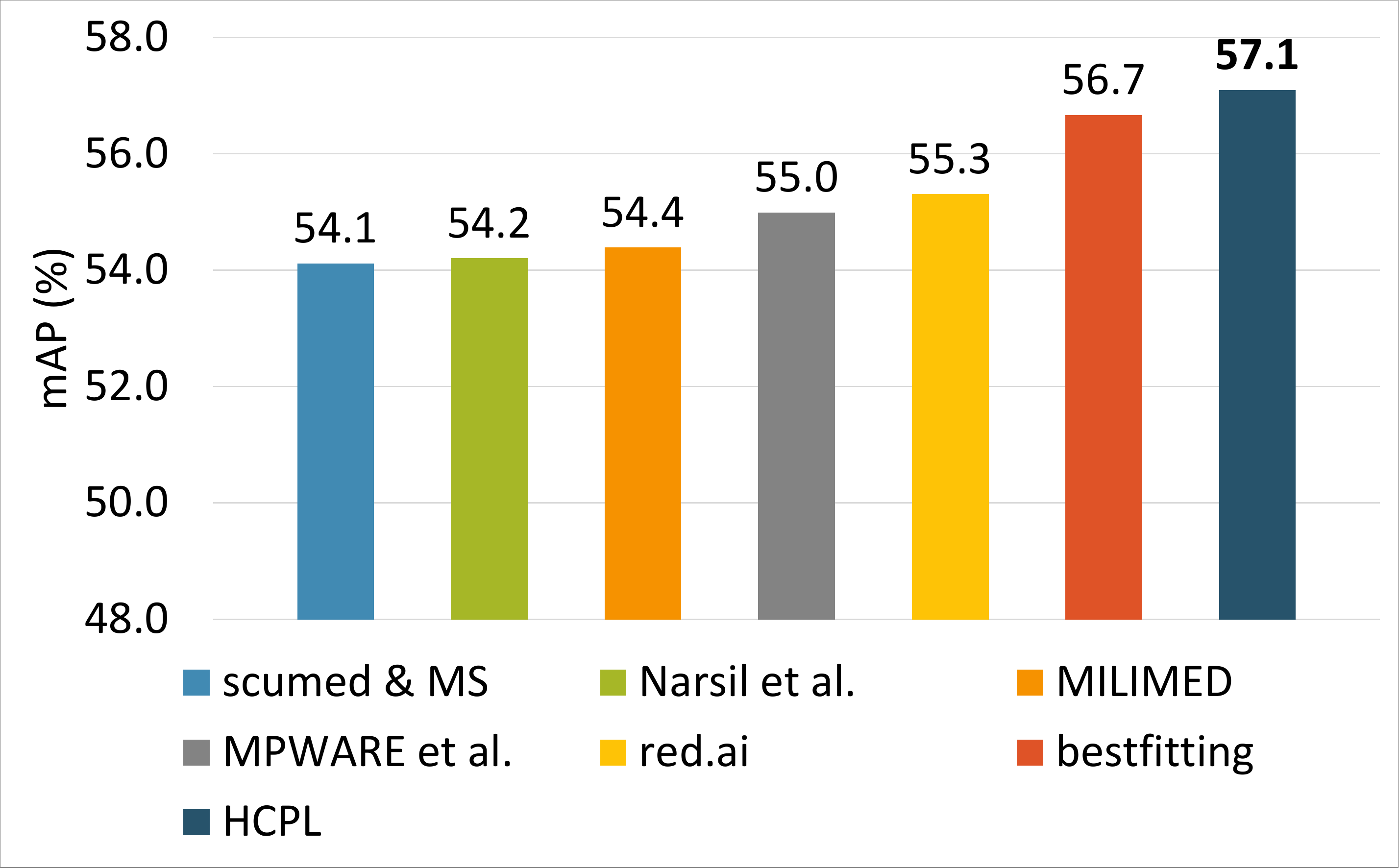}
         \vspace{-0.5cm}
         \label{fig:sota}
     \end{subfigure}
        \hfill
     \begin{subfigure}[b]{1\textwidth}
         \centering
         \caption{}
         \includegraphics[width=\textwidth]{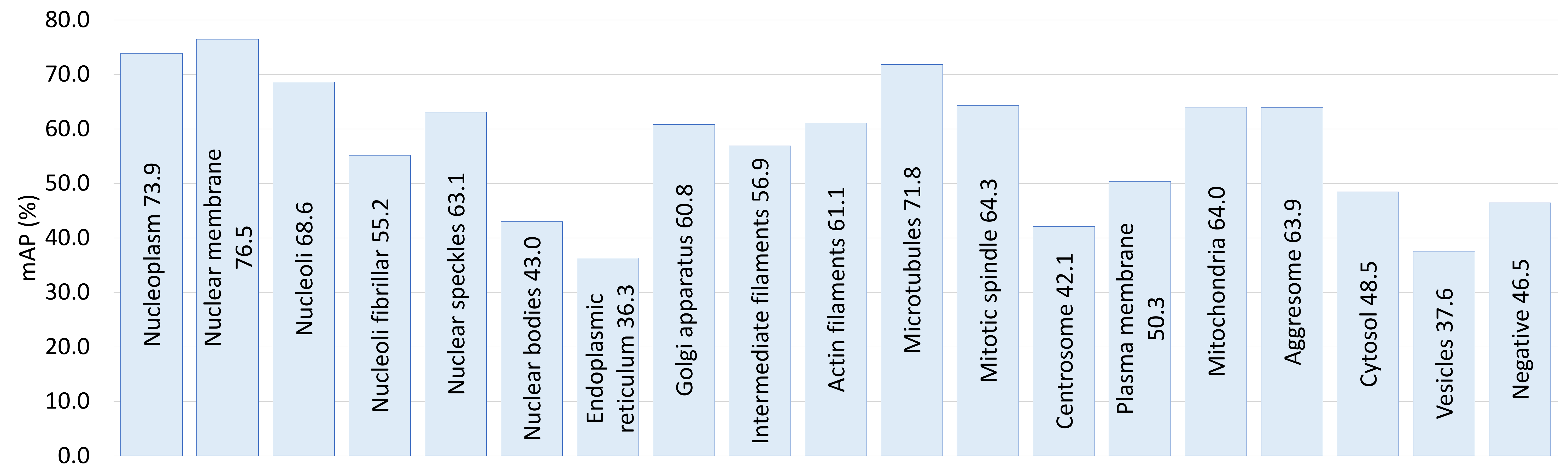}
         \label{fig:classwise}
     \end{subfigure}
        \caption{\textbf{Evaluation of the HCPL system}. \textbf{a}, The impact of diversity-based ensembling of DNNs. \textbf{b}, Comparison of our DSA with Kaggle multi-head models. \textbf{c}, Comparison of our CLA with existing cell-level models. \textbf{d}, Comparison of our final system with state-of-the-art ensembled systems. \textbf{e}, The performance of our system for different label classes. Performance of the two rarest classes is relatively good: mitotic spindle 64.3\% and aggresome 63.9\%. The nuclear bodies class achieves only 43.0\% as it is often mistakenly classified as nucleoplasm. Furthermore, there is a significant amount of confusion between cytosol, endoplasmic reticulum and plasma membrane classes resulting in relatively poor performance.}
        \label{fig:three graphs}
\end{figure*}

\noindent\newline\textbf{Robust protein localisation using diversity-based ensembling.}
Experimental results of our models with eighteen base DNNs show that the maximum performance that a single model can achieve is 55.8\% mAP (Figure \ref{fig:ens}). Classification accuracy can be significantly improved by optimal ensembling DNNs (classifiers) outputs. However, meticulous selection of diverse classifiers is required to achieve this improvement. A set of classifiers are considered diverse if they perform well on different examples or classes.

The diversity inherent in the DNNs can be visualised with a correlation matrix of probabilities produced by each network (Figure \ref{fig:model_corr}). 
Here, we observe a block diagonal structure in the correlation matrix (green diagonal blocks), demonstrating that DNNs of similar architectures (EfficientNets, NFNets, ResNets, ResNests, DenseNets and Transformers) have strong correlations. The selection of the final network set (Figure \ref{fig:inference}) is accomplished by selecting the best performing DNN, on the Kaggle public leaderboard (Methods), from each class of architectures.

Figure \ref{fig:inference} demonstrates the HCPL system, which employs nine diverse networks. The first phase is to extract individual cells from each image using HPA Cell Segmentator. Next, the cells are forwarded to the individual DSA, Cell-level networks to compute the predictions. The outputs from these diverse networks are hierarchically aggregated to compute the probabilities of all classes. Simultaneously, the cells are passed to the VID to compute the cell visual integrity weighting. Finally, the probabilities are multiplied by the visual integrity weights to generate the final vector of class probabilities.

\noindent\newline\textbf{Quantifying HCPL performance and identifying its essential components.}
We conduct ablation studies to evaluate  strengths of different models and the improvements brought by diversity-driven multi-DNN ensembling.

Figure \ref{fig:ens} shows that the individual DNNs are not able to improve beyond 55.8\% mAP. The mAP is improved to 56.6\% by ensembling four Dual-stream architectures with bases EfficientNet-B4 \cite{pmlr} (54.7\%), NFNet-ECA \cite{nfnet} (54.8\%), Regnety-32 \cite{RegNet} (55.1\%) and DenseNet161 \cite{dense} (54.6\%). We can achieve a mAP of 56.7\% by ensembling the predictions from CLAs with bases EfficientNet-B4 (55.6\%), SwinT \cite{swint} (54.9\%), ResNet200d \cite{Xie2019} (55.8\%), ResNest50d (55.4\%) and CLH (55.2\%). Importantly, the ensembling of all DSA and Cell-level networks obtain a state-of-the-art classification score of 57.1\%. 

Figure \ref{fig:classwise} demonstrates the performance of our system in each of the nineteen classes. The performance depends on several variables such as the difficulty of recognising different localisation patterns, number of training samples and extreme visual variations. Despite a lower number of training samples, the mAP is higher for aggresome (63.9\%) than plasma membrane (50.3\%) because aggresome has visually distinguishable features, whereas plasma membrane is often confused with cytosol. The endoplasmic reticulum has the lowest mAP of 36.3\%; it is also confused with cytosol. Centrosome also has a low mAP of 42.1\% due to its proximity to the nucleus, causing it to often appear within the nucleus; a consequence of 2D projection of a 3D structure. This means it can be confused with nucleoli or nuclear bodies. Despite the extreme rarity of the mitotic spindle in training samples, our system achieves a good score of 64.3\%. 

\begin{figure*}
     \centering
             \vspace{-0.3cm}
     \begin{subfigure}[b]{1\textwidth}
         \centering     
         \caption{}
         \vspace{-0.3cm}
         \includegraphics[width=\textwidth]{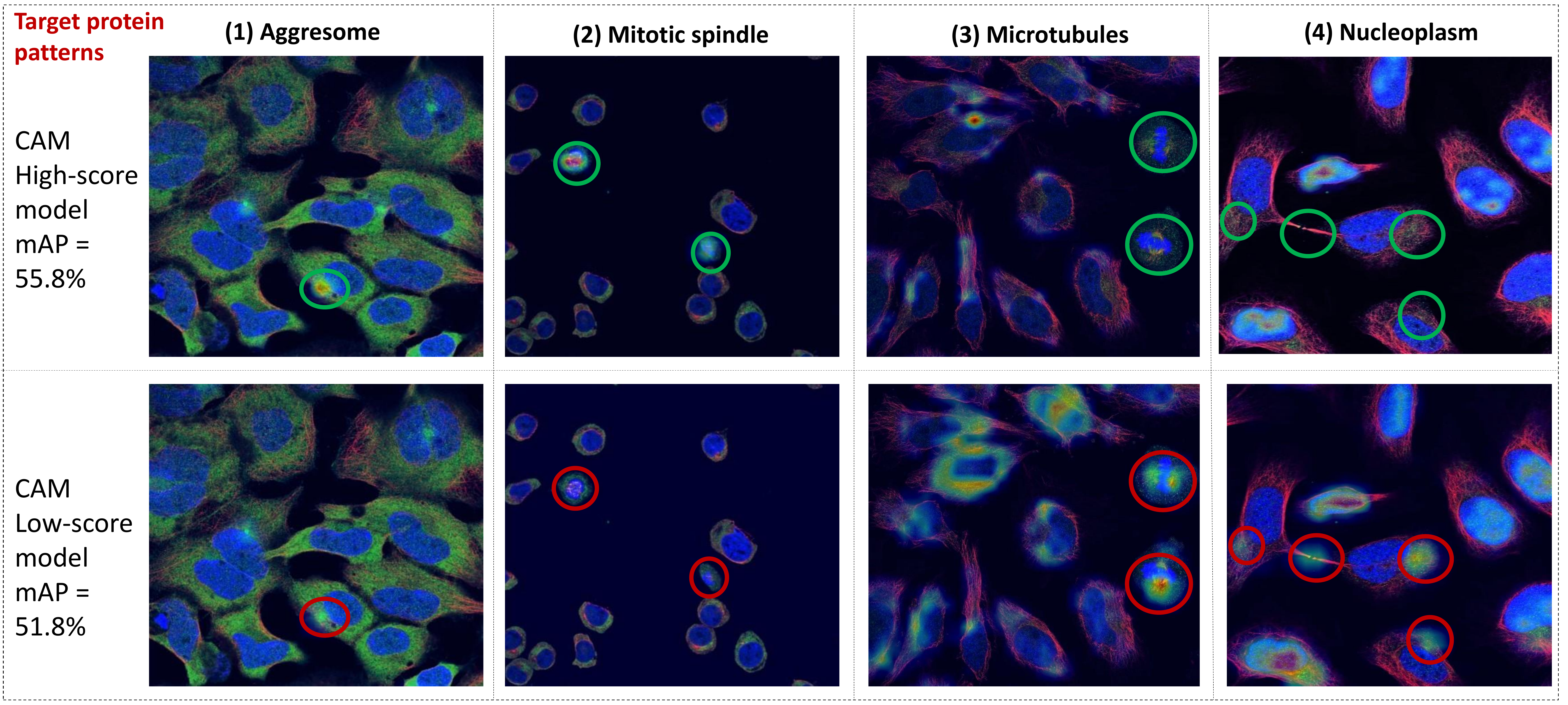}
        \label{fig:CAMs}
         \vspace{-0.5cm}
     \end{subfigure}
     \hfill
     \begin{subfigure}[b]{1\textwidth}
         \centering
        \caption{}
        \vspace{-0.25cm}
         \includegraphics[width=\textwidth]{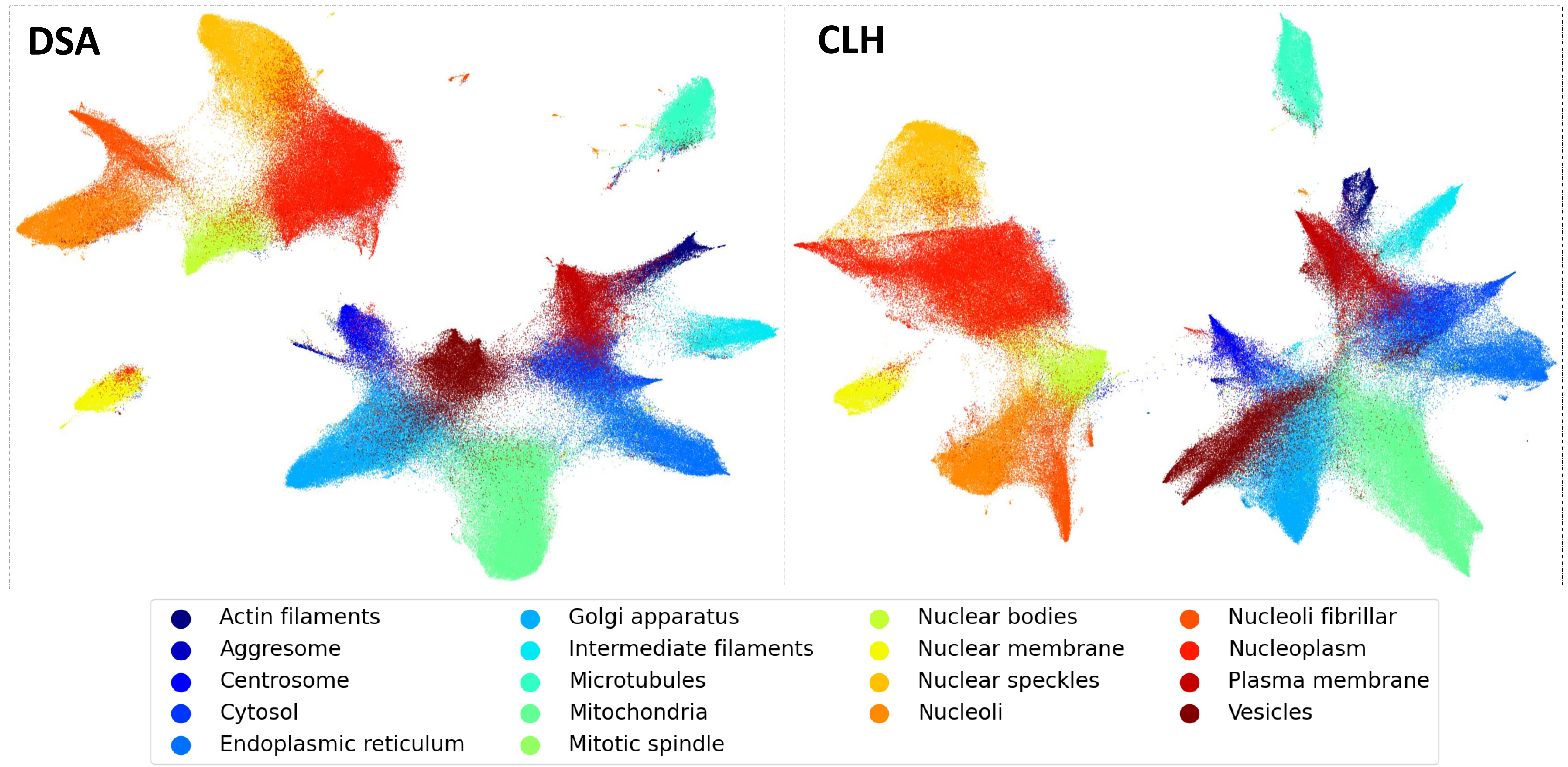}
        \label{fig:UMAP}
     \end{subfigure}
     \caption{\textbf{Visual interpretations of learned deep features}. \textbf{a}, CAMs of the high-scoring CLA trained on CRA improved labels, and a low-scoring CLA trained on original weak labels. (1) The red circle shows where the aggresome staining pattern is not captured by the low-scoring DNN, whereas the green circle depicts the high-scoring model correctly localises it. (2) For the mitotic spindle staining patterns, CAMs for the high-scoring CLA highlight relevant cellular regions indicated by green circles. (3) The microtubule staining overlaps well with the CAM for the high-scoring model. In contrast, the low-scoring model uses visual features of the mitotic spindle class to assign a high probability for the microtubule class, as indicated by red circles. (4) The CAMs for the nucleoplasm staining demonstrate biologically relevant attention for the high-scoring model only. \textbf{b}, Visualisation of the deep features for DSA and CLH DNNs from the UMAP dimensionality reduction method. High-dimensional data is projected onto a 2D plane such that local structures in the original space are captured, whilst simultaneously retaining the global structure of the data. Different subcellular locations are effectively clustered by deep features in each model.}
\end{figure*}

\noindent\newline\textbf{Benchmarking HCPL system.} 
Since no published works on single-cell classification exist, a starting reference point is the best performing image-level classification DNN (bestfitting, DenseNet-based) \cite{kagglenature}. This model was fine-tuned on the HPA single-cell classification dataset. The resulting performance is 33\% mAP which shows that image-level classifiers are not directly applicable to single-cell classification. A more insightful comparison can be obtained by evaluating the performance of HCPL against the methods developed in the recent Kaggle competition \cite{HPAresults}. 

We first compare the performances of single models without ensembling. From Figure \ref{fig:dual-sota}, we observe that the proposed DSA outperforms Kaggle multi-head models including Fair Cell Activation Network (FCAN) \cite{FCAN}, Duo-branch \cite{DBCM}, modified Puzzle-CAM \cite{pizzlecam} and PfNet \cite{seventh}. Among cell-level architectures, Figure \ref{fig:cell-sota} demonstrates that our CLA achieves the best classification performance of 55.8\% outperforming Cell model InceptionV3 \cite{FCAN}, Cell model SwinT \cite{FCAN}, Data-centric CNN \cite{datacentric}, Gridify and Gapmask \cite{sixth} and Bag-of-Cells \cite{BCM}.

When ensemble-based architectures are considered (Figure \ref{fig:sota}), our HCPL system achieves the highest classification accuracy of 57.1\% mAP. The second-best result (56.7\%, bestfitting, \cite{FCAN}) was trained with antibody information; such an approach is generally discouraged as it is known to generalise poorly on unseen cell lines or/and new staining markers. A previous study \cite{kagglenature} reported that gains achieved using this information might be due to the exploitation of data leakage between the training and testing examples via the antibody identifiers and cell-line information specific to the competition data. The next best approach \cite{DBCM} that did not utilise antibody data achieved significantly lower performance of 55.3\%.

\noindent\newline\textbf{The ability of HCPL to interpret biological information.}
We use the Grad-CAM method \cite{gradcam} to identify the parts of an input image that  impact the classification score. These attention regions will vary for each image and location label. This highlights regions contributing features that support predictions of the target label, helping us understand whether localisation predictions are biologically meaningful (by comparing to the ground truth staining patterns).

Figure \ref{fig:CAMs} shows CAM regions for challenging patterns, such as aggresome, mitotic spindle, microtubules, and nucleoplasm. It compares the low-scoring CLA (trained on weak cell labels) and high-scoring CLA (trained using strong labels obtained from the CRA). We can observe that the staining patterns for these difficult classes overlap well with the corresponding CAM attention regions of the high-scoring model, hence confirming that our high-scoring CLA focuses on biologically relevant cell regions.

\noindent\newline\textbf{Deep features visualisation using UMAP.}
To investigate the ability of a DNN to distinguish 
subcellular structures, we visualise the high-dimensional feature vector extracted from the penultimate layer using uniform manifold approximation and projection for dimension reduction (UMAP) \cite{UMAP}. The visualisation results for two different DNNs (DSA and CLH) are presented in Figure \ref{fig:UMAP}. Here, each point on the plot represents a single cell. Only cells with single labels are chosen, coloured by their respective label. 

Firstly, we observe that the DNNs cluster the majority of cells from the same class together, which shows their underlying discriminative power. Secondly, we note the presence of two larger clusters at a global level: one for nuclear sub-compartments (nucleoplasm, nuclear membrane, nucleoli, nucleoli fibrillar center, nuclear speckles and nuclear bodies) and another for locations outside the nucleus (e.g. cytosol, mitochondria). Thirdly, instances of some classes are located within clusters belonging to other classes, which links to the relative decrease in performance (Figure \ref{fig:classwise}), for example: cytosol partially overlaps with endoplasmic reticulum and plasma membrane, nuclear bodies with nucleoplasm and vesicles with Golgi apparatus. Lastly, the different amount of overlap observed in the clusters generated by DSA and CLH confirms the diversity and complementarity in the deep features.

\section*{Discussion} \label{discussion}
Our core contributions include a novel DSA architecture for improved image and cell information fusion and two novel Cell-level architectures (CLA \& CLH) to capture the wide variability between cells. Further, we introduce an ``AI-trains-AI'' approach with DNN-powered re-labelling and automatic cell visual integrity weighting and propose an effective strategy to optimally ensemble diverse DNNs. The evaluation shows that HCPL outperforms existing approaches and is expected to generalise well to unseen cell lines and proteins.
Further improvements to the state-of-the-art could be achieved by obtaining more examples of rare localisation classes, accurate cell labels, and feedback from human experts.

HCPL narrows the performance gap between AI methods and human experts and provides a toolbox of methods to tackle the challenges of single-cell protein localisation successfully. This will help accelerate the characterisation of unknown proteins and our understanding of cellular function and biology to advance our knowledge of disease-related phenotypes and drug discovery. 

\section*{Methods}
\label{sec:methods}
\noindent\newline\textbf{HPA Dataset.}\label{Dataset}
Our work uses the HPA dataset provided by the ``Human Protein Atlas - Single Cell Classification'' Kaggle challenge \cite{HPA}. This consists of images from the freely accessible Human Protein Atlas project. In particular, images from the subcellular section of the HPA were used. A total of 97K images (internal and external dataset) were made available for training purposes whilst an additional 2270 unseen images were retained by the challenge organisers for testing purposes, further split into 559 images for the public leaderboard and 1711 images for the private leaderboard. Each of the provided images contains multiple cells and consists of four channels, displayed in red, green, blue, and yellow. The task considered in this paper is to localise the protein of interest (green channel) in 18 possible subcellular organelles in each cell in an image. An additional ``negative'' class is added for negative staining and unspecific patterns. A cell can have multiple labels
(specifically, we work with 19 localisation classes with up to 6 protein locations per cell). The 19 labels and their names are shown in Figure \ref{fig:challenges}. Our DNNs are trained on approximately 1.2 Million cell images segmented from HPA images using the HPA Cell Segmentator.

\noindent\newline\textbf{Evaluation metric.} \label{eval-measure}
To ensure a fair comparison of results, all experiments were evaluated by computing mean Average Precision (mAP) \cite{map}. The mean value was calculated over the 19 segmentable classes ($C$) of the challenge with a mask-to-mask IoU $>$ 0.6 as described below:

\begin{equation}
    \mathbf{mAP} = \frac{1}{C}\sum^{C}_i Prec_{i}
\end{equation}

where $Prec_i$ is Precision for class $i$ which is calculated according to the two-stage method described in article \cite{everingham}.

All mAP scores are reported based on the Kaggle private leaderboard.

\noindent\newline\textbf{HPA Cell Segmentator (HCS).}
The HCS \cite{segmentator} segments input images into individual cell instances for multi-label classification. Since the procedure of the HCS software provided by organisers occupied 60\% of the total permitted processing time (9 hours for the entire system's inference), improving the efficiency of the segmentation algorithm is important. The algorithm consists of three main stages: i) prediction of the nuclei; ii) general cell prediction and iii) post processing procedure. However, cell segmentation training data was not publicly available, and so gains could only be obtained by modifying the post-processing procedure rather than training a new model. In the segmentation architecture, the nuclei and cell maps are first obtained via the corresponding predictor U-Net \cite{unet}. The next step is the post-processing of the outputs. To improve the efficiency of HCS, we introduce down-scaling and up-scaling blocks at the start and the end of the process respectively. Since post-processing is largely based on various morphological operations, its complexity is proportional to the product of the processed image dimensions. Therefore, reducing the spatial dimensions by 50\% resulted in a speedup of a factor of at least 2. Another effect of the reduced dimensionality was a significantly lowered amount of high-frequency noise. This allowed simplification of the pipeline by removing extra processing in two fine-tuning blocks and in the "Segmentation and gradual object removal" block. These changes resulted in a 2x speed-up. However, such speed improvement resulted in a deterioration in system accuracy of around 0.2\% mAP.

\noindent\newline\textbf{The Weibull activation layer.}
\label{methods:weibull}
From each image, $N$ cells are selected, resized and flattened as a batch (typically $N=20$). Let $\mathbf{X} \in \mathbb{R}^{A\times B \times 4}$ denote a cell image of resolution $A \times B$. Each cell $\mathbf{X}$, is processed by a base DNN (e.g. EfficientNet \cite{pmlr}), which embeds an input into the space of compact deep features. The output tensor of the final convolutional layer, denoted as $\mathbf{R} \in \mathbb{R}^{W\times H \times D}$, is forwarded to a learnable activation layer \cite{WAP}, where $W$ and $H$ are the width and height of the feature map and $D$ is the feature dimensionality. The Weibull activation layer is aimed at maximising the Signal-to-Noise ratio (SNR) of the last convolutional feature map by applying the Weibull function to the tensor $\mathbf{R}$, the output tensor of the final convolutional layer. Each element of the tensor $\mathbf{R}$ is transformed by the Weibull function resulting in the output tensor $\mathbf{T} \in \mathbb{R}^{W\times H \times D}$ (where $0 \leq i \lt W \times H \times D$): 

\begin{equation}
\mathbf{T}_i = \left(\frac{ \mathbf{R}_i}{\lambda}\right)^{\zeta-1}\exp(-(\mathbf{R}_i/\gamma)^\eta).
\end{equation}

The learnable parameters of the activation layer are $\lambda, \zeta, \gamma$, and $\eta$. The output of the activation layer is fed to the Global Average Pooling (GAP) layer, denoted as $P(\mathbf{T})$, to compute the global vector $\mathbf{S}$:
\begin{equation}
    \mathbf{S} = \left(\frac{1}{WH}\sum^W_i\sum^H_j T_{ijk} \right)_{k = 1}^D
\end{equation}
Each element ($s\in \mathbf{S}$) of the tensor $\mathbf{S}$ is power-normalised to balance the non-linear scaling of the Weibull function. The power normalisation function is represented as $\delta:\mathbb{R}^D\rightarrow\mathbb{R}^D$, with the rule:
\begin{equation}
    \delta(s) = \alpha s_1^{\beta},\alpha s_2^{\beta} ,..., \alpha s_D^{\beta}
\end{equation}
where $\alpha$, $\beta$ are learnable scaling parameters.

\noindent\newline\textbf{DSA, CLH and CLA training and inference configurations}
 The DSA comprises a baseline DNN followed by the dual stream network.
From each image, $N$ cells are selected, resized and flattened as a batch (typically $N=20$). Let $\mathbf{X} \in \mathbb{R}^{A\times B \times 4}$ denote a cell image of resolution $A \times B$. Each cell $\mathbf{X}$, is processed by a base DNN, which embeds an input into the space of compact deep features. The output tensor of the final convolutional layer, denoted as $\mathbf{R} \in \mathbb{R}^{W\times H \times D}$, is forwarded to a learnable Weibull activation layer \cite{WAP}, where $W$ and $H$ are the width and height of the feature map and $D$ is the feature dimensionality. The output of the activation layer is forwarded to a Global Average-Pooling (GAP) layer and power-normalisation layer to generate global descriptors, which are then passed to the image stream and cell stream. The image stream applies Global Max-Pooling to a bag of $N$ cell descriptors originating from a single image to generate a unified image representation $V$, which is then passed to a fully connected layer ${FC_1}$ and Softmax to generate an image-level prediction. The cell stream takes $N$ cell descriptors as an input and outputs the predictions for each cell using a fully-connected layer ${FC_2}$ and Softmax. The predictions from the image stream are passed to classification loss layer. The loss layer computes the weighted Binary Cross-Entropy loss ($L_1$) between the image label and bag-prediction. Similarly, the cell stream weighted Binary Cross-Entropy loss $L_2$ is calculated between cell predictions and cell labels. The final loss ($L_f$) is the weighted sum of cell stream loss and image stream loss $L_f=W_1\times L_1 + W_2\times \L_2$. For the cell stream, the labels are weak and we therefore intuitively assign a much lower weight to cell stream loss ($W_2 = 0.2\times W_1$). The DSA is trained using an Adam optimiser and cosine annealing learning rate scheduler.  

The Cell-level Hybrid takes cell images as an input and outputs convolutional features denoted as $\mathbf{R} \in \mathbb{R}^{W\times H \times D}$. The features are aggregated using Global Average Pooling (GAP) layer and forwarded to classification module (fully-connected layer and Softmax). 

In Cell-level Actnet, the convolutions features extracted from cell images are passed to learnable Weibull activation pooling. The transformed features are aggregated using GAP and power-normalisation layers and forwarded to the classification module.

The training of CLH and CLA is performed using weighted Binary Cross-Entropy loss, Focal loss, Adam optimiser, and a cosine annealing scheduler with initial learning rate $2e^{-4}$.

We applied data augmentation in the form of random cropping, flipping, shifting, rotation, scaling and cutout to train all models.

\noindent\newline\textbf{Visual Integrity Detector system.}
For training of first VID module, we first compute the eight most representative features from each cell in the training dataset: bounding box height, width, aspect ratio, area, mask area, mask perimeter, the value of the largest dimension and a binary feature that is based on the pixel intensity and the ratio of blue and green to the total number of pixels. The training dataset contains 10K cells hand-labelled as either 'good', i.e. most of a cell is clearly visible or 'bad', i.e. a cell is damaged and not suitable for further processing. Note, this dataset with an extended set of properties and reference segmented cells is made publicly available \cite{BCDdataset}. We then train the XGBoost classifier on cell features using a five fold cross-validation strategy.

The second module consists of a base EfficientNet-B2 with a fully connected layer to output predictions for four classes. The data to train the DNN is generated by randomly cropping out some area on the border of the cell. If the cropped area is less than 30\% of the original cell, that image belongs to class 1. Similarly, if the cropped area is between 30\% to 50\%, 50\% to 80\% and 80\% to 100\% then that image is assigned to classes 2, 3 and 4 respectively. The network takes cropped cell tiles as an input and outputs probabilities of the four classes. The training is performed using the cross-entropy loss function. At inference time, each cell is forwarded to a trained EfficientNet-B2, and the probability for each class is obtained.

\bibliography{ref.bib} 

\subsubsection*{Author contributions}
All authors contributed equally to the manuscript.

\subsubsection*{Competing interests}
The authors declare no competing interests.


\end{document}